\documentclass[10pt,journal,compsoc]{IEEEtran}
\usepackage{color}
\usepackage{amssymb}
\usepackage{amsmath}
\usepackage{ragged2e}
\usepackage{multirow}

\ifCLASSOPTIONcompsoc
  \usepackage[nocompress]{cite}
\else
  \usepackage{cite}
\fi

\ifCLASSINFOpdf
  \usepackage[pdftex]{graphicx}
\else
  \usepackage[dvips]{graphicx}
\fi

\usepackage{url}

\begin{document}

\title{Exploration of Class Center for Fine-Grained Visual Classification}

\author{Hang~Yao,
		Qiguang~Miao,~\IEEEmembership{Senior Member,~IEEE,}
        Peipei~Zhao,
        Chaoneng~Li,~\IEEEmembership{Student Member,~IEEE,}
        Xin~Li,
        Guanwen~Feng,
        and~Ruyi~Liu
\IEEEcompsocitemizethanks{\IEEEcompsocthanksitem Peipei Zhao is the corresponding author\protect\\
E-mail: zhpp2023@xidian.edu.cn

\IEEEcompsocthanksitem Hang Yao, Qiguang Miao, Peipei Zhao, Chaoneng Li, Guanwen Feng and Ruyi Liu are affiliated with the School of Computer Science and Technology, Xidian University, Xi'an, Shaanxi 710071, China, Xi'an Key Laboratory of Big Data and Intelligent Vision, Xi'an, Shaanxi 710071, China, Key Laboratory of Collaborative Intelligence Systems, Ministry of Education, Xidian University, Xi'an 710071, China.\protect\\
Xin Li is affiliated with the School of Mechanical Engineering, Yanshan University, Qinhuangdao 066004, China.}}

\IEEEtitleabstractindextext{
\begin{abstract}
\justifying
Different from large-scale classification tasks, fine-grained visual classification is a challenging task due to two critical problems: 1) evident intra-class variances and subtle inter-class differences, and 2) overfitting owing to fewer training samples in datasets. Most existing methods extract key features to reduce intra-class variances, but pay no attention to subtle inter-class differences in fine-grained visual classification. To address this issue, we propose a loss function named exploration of class center, which consists of a multiple class-center constraint and a class-center label generation. This loss function fully utilizes the information of the class center from the perspective of features and labels. From the feature perspective, the multiple class-center constraint pulls samples closer to the target class center, and pushes samples away from the most similar nontarget class center. Thus, the constraint reduces intra-class variances and enlarges inter-class differences. From the label perspective, the class-center label generation utilizes class-center distributions to generate soft labels to alleviate overfitting. Our method can be easily integrated with existing fine-grained visual classification approaches as a loss function, to further boost excellent performance with only slight training costs. Extensive experiments are conducted to demonstrate consistent improvements achieved by our method on four widely-used fine-grained visual classification datasets. In particular, our method achieves state-of-the-art performance on the FGVC-Aircraft and CUB-200-2011 datasets.
\end{abstract}

}

\IEEEpubid{\begin{minipage}{\textwidth}\ \centering
		Copyright © 2024 IEEE. Personal use of this material is permitted. \\ However, permission to use this material for any other purposes must be obtained from the IEEE by sending an email to pubs-permissions@ieee.org.  DOI:10.1109/TCSVT.2024.3406443
\end{minipage}}
\maketitle

\section{Introduction}

\IEEEPARstart{A}{s} an extension of generic image classification (e.g. ImageNet classification \cite{1}), fine-grained visual classification (FGVC) aims to recognize different subcategories belonging to a basic-level category (e.g., birds, cars, and aircraft). In FGVC tasks, samples from the same class show evident differences in posture of objects, lighting and backgrounds. Moreover, because of their similar appearances, samples from different classes are easily confused. Thus, FGVC exhibits obvious intra-class variances and subtle inter-class differences. Moreover, there are fewer training samples in each category in datasets, which leads to overfitting when large-scale deep neural networks are trained. Therefore, FGVC is a challenging task.

\begin{figure*}
	\centering
	\label{fig1}
	\includegraphics[width=1.0\linewidth]{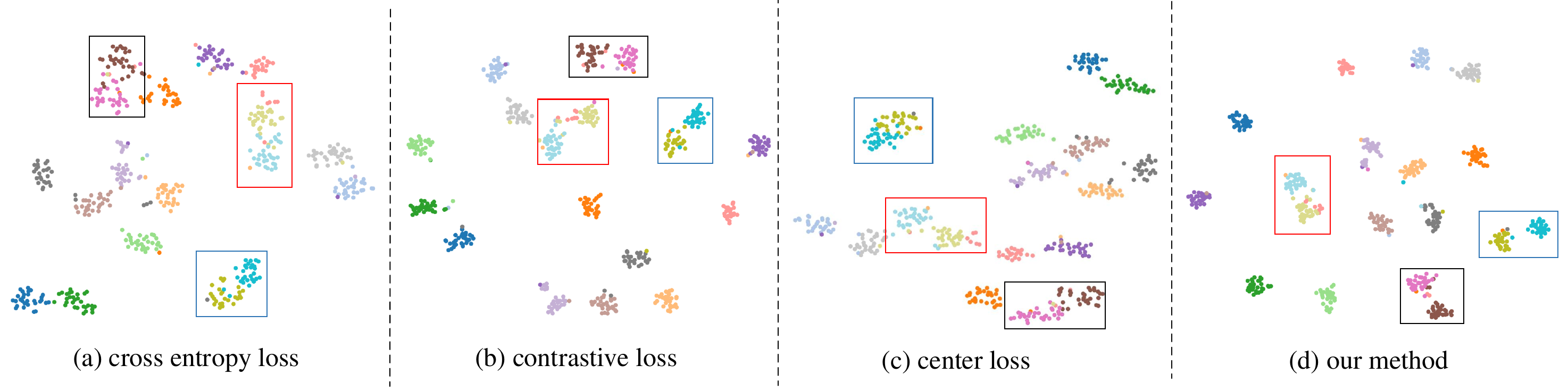}
	\caption{The t-SNE results of (a) cross entropy loss, (b) contrastive loss, (c) center loss and (d) our method on 18 categories of warblers. The improvements in (b) contrastive loss and (c) center loss are limited by inter-class differences and intra-class variances, such as classes masked in boxes. Compared with other methods, (d) our method compresses samples of the same class into a compact cluster and significantly enlarges the margins between different clusters, especially for the classes masked in the boxes. Thus, our method effectively reduces intra-class variances and enlarges inter-class differences.}
\end{figure*}

Recent FGVC methods design complex networks to focus on object areas to ignore cluttered backgrounds\cite{4,7,39,76} or extract the features of parts to reduce the impact of posture\cite{2,3,5,6,8,85}. Thus, these FGVC methods significantly reduce intra-class variances. However, most of these methods rely only on cross entropy loss to obtain classification boundaries, which is insufficient to handle inter-class differences. In addition, some common visual classification methods introduce class center as representations of whole classes, and reduce the distances between samples and class centers to reduce intra-class variances\cite{45,11,46,47}. In addition, Zhang et al. proposed a feature aggregation scheme to resist intra-class variances\cite{83}. However, these methods do not consider inter-class differences in the FGVC, which limits further improvement in model performance. For example, in \textcolor{blue}{Fig.\ref{fig1}} (a), there are no clear classification boundaries between some closer class clusters that are masked with boxes. The same issue occurs in the t-SNE results of center loss\cite{45} in \textcolor{blue}{Fig.\ref{fig1}} (c). The class clusters masked by the blue box are even closer than those in \textcolor{blue}{Fig.\ref{fig1}} (a). Thus, we should further consider inter-class differences in FGVC.

To simultaneously handle inter-class differences and intra-class variances, some common visual classification methods utilize contrastive learning to constrain the feature distances of positive and negative sample pairs\cite{10,74,75,79,84}. However, these methods are not suitable for FGVC for two reasons. First, the optimization direction of each positive or negative pair is not consistent, which limit improvement of the FGVC. For example, consider three samples, A1, A2 and A3 which belong to class A, and a sample B1 that belongs to class B. Contrastive learning attempts to optimize the distance between a positive sample pair (A1 and A2) and the distance between a negative sample pair (A3 and B1). A1 and A2 are pulled closer, and A3 is pushed away from B1. However, the distance between A1 and A3 may be greater, and A1 and B1 may be closer. In this case, the optimization directions of A1 and A3 are inconsistent. This method can neither guarantee that A1, A2 and A3 are clustered together nor ensure that the distance between A1 and B1 is larger. If we consider only optimization between samples, optimizing one sample pair may have a negative effect on other samples. As shown by the t-SNE results on 18 categories of warblers in \textcolor{blue}{Fig.\ref{fig1}} (b), the improvement in the contrastive loss is limited. Class clusters do not pull samples from the same class into compact clusters (such as class clusters masked by blue and black boxes), and distances between some class clusters are not significantly widened (such as class clusters masked by blue and red boxes). Second, in fine-grained image classification, one category is usually very similar to one or two other categories. Therefore only the differences between these similar categories need to be expanded. Because of the absence of prior knowledge concerning inter-class similarity, the above contrastive learning methods treat different nontarget classes equally. Therefore, these methods do not effectively expand the differences between similar categories and are not suitable for handling inter-class differences in the FGVC. How to compress intra-class variances while effectively expanding the distance between similar classes is a problem that needs to be solved.

\begin{figure*}
	\centering
	\includegraphics[width=1.0\linewidth]{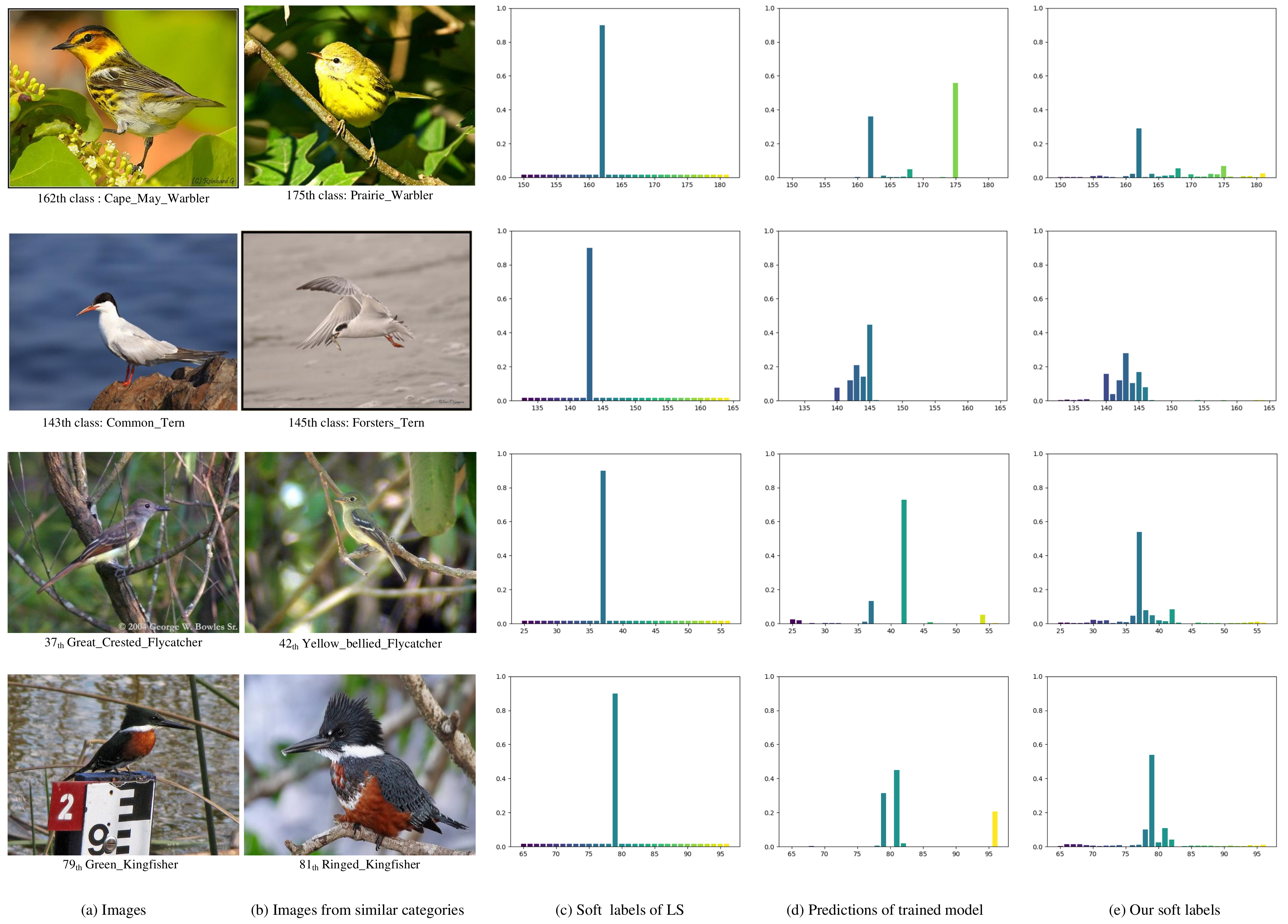}
	\caption{There are four examples of soft labels, which correspond to the images in the first column. The columns from left to right show (a) images, (b) images from similar categories of (a), (c)smooth labels of LS, (d) predictions of the trained model and (e) our soft labels from CLG. Columns (a) and (b) are visually similar samples but belong to different categories. In Column (c), LS assigns the same confidence to all nontarget classes. Such soft labels do not reflect relationships between classes. The confidence of nontarget classes should be positively related to the similarity between the target class and nontarget classes. Other methods utilize the predictions of trained models as soft labels. However, the predictions may be incorrect, as shown in Column (d). Some samples can easily be predicted as similar nontarget classes, whose samples are shown in Column (b). Different from smooth labels of LS and predictions of trained model, our labels in Column (e) reflect the similarity between classes and ensure correct labelling.}\label{fig2}
\end{figure*}

In addition, there is also the issue of overfitting in FGVC. Soft labels are considered an effective way to address overfitting\cite{72}. Label smoothing (LS) uniformly reassigns partial confidence of the target class to the nontarget categories to generate soft labels \cite{18}. Nevertheless, uniform confidence in nontarget categories ignores rich correlations between fine-grained categories \cite{11,17}. Compared with other nontarget classes, more similar nontarget classes should be assigned more confidence. For instance, \textcolor{blue}{Fig.\ref{fig2}} displays some images of birds in (a) and (b) shows samples from the classes that are similar to (a). The soft labels of LS in \textcolor{blue}{Fig.\ref{fig2}} (c) do not reflect the similarity between (a) and (b). In the soft labels of (a), classes of (b) should be assigned more confidence than other nontarget classes. Label refinery (LR) and guided label refinery (GLR) use the predictions of the trained model as soft labels to reflect similarity\cite{20, 73}. However, because similar categories are difficult to distinguish, model predictions are likely inaccurate and may generate incorrect soft labels. As shown in \textcolor{blue}{Fig.\ref{fig2}} (d), the trained model incorrectly predicts the images in (a) as classes in (b). Thus, the problem of how to generate reliable and reasonable soft labels remains.

Motivated by the above issue, we propose a simple but effective method named exploration of class center (ECC), which fully mines the information of class center from the perspectives of features and labels. Our ECC consists of 1) a multiple class-center constraint (MCC) and 2) a class-center label generation (CLG). From the feature perspective, MCC constructs class-center features to provide overall representations of the classes. The feature distance between the sample and target class center is constrained to compress intra-class variances. Then the cosine similarity between class-center features is calculated as the similarity between classes. According to the similarity, the most similar nontarget class can be searched, and inter-class differences are enlarged by constraining the feature distance between the sample feature and class-center feature of the most similar nontarget class. By constraining the distances between samples and class centers (the target class and the most similar nontarget class center), we can pull samples close to the target class centers, and push samples away from the most similar nontarget class centers. Thus, we can guarantee consistent sample optimization directions. We can also specifically address the differences between samples and the most similar nontarget categories with similarity between class center features. \textcolor{blue}{Fig.\ref{fig1}} (d) shows that compared with other methods, our method results in class clusters that are tightly gathered and larger distances between different class clusters. From the label perspective, the CLG employs class-center distributions to generate reliable soft labels, as shown in \textcolor{blue}{Fig.\ref{fig2}} (e), to alleviate overfitting and further introduce correlations between classes. Finally, ECC and cross entropy (CE) loss are combined to optimize the model. In addition
, different from existing methods based on class center, a novel strategy for updating class center is proposed to update class center more stably. Our method addresses FGVC tasks without complex structures or training strategies, thus, allowing our approach to be easily integrated into other FGVC methods to further boost performance. Extensive experiments are conducted on FGVC-Aircrafts (AIR) \cite{21}, CUB-200-2011 (CUB) \cite{22}, Stanford Cars (CAR) \cite{23} and NABirds (NAB) \cite{91}. The results prove effectiveness of our proposed approach.

\section{Related Work}
\subsection{Fine-grained Image Classification}
Benefiting from the development of deep learning \cite{24,25}, there has been significant progress in existing FGVC research in recent years \cite{80,81,82}. FGVC methods can be divided into two categories based on whether they employ extra manual annotations: strongly supervised methods and weakly supervised methods. Strongly supervised methods require manually labelled bounding boxes or part annotations, with which informative key parts can be located for extracting discriminative part features \cite{33,34,35,36,37}. Finally, key part features and object features are integrated for classification. However, these additional manual annotations require extensive expert knowledge. There is limited feasibility and scalability in real-world applications. Therefore, weakly supervised methods without manual annotation have attracted much attention from researchers \cite{81,82,86,87,94}. M2DRL\cite{93} learns multigranular discriminative region attention and multiscale region-based feature representation for more accurate object region positioning and category recognition. DME-Net\cite{81} introduces a multitasking framework for the low-resolution fine-grained image recognition task, that aims to capture reliable object descriptions from macro- and microperspectives, respectively. SIM-Trans\cite{92} incorporates object structure information into transformer to enhance discriminative representation learning to contain both appearance information and structure information. SIA-Net\cite{82} extracts the low-level image details under the guidance of accurate semantics and makes the details spatially correspond to high-level semantics with complementary content. AA-Trans\cite{85} acquires discriminative parts of the image precisely to better capture local fine-grained information. MA-CNN\cite{2} locates part localization with proposed channel grouping layers in a weakly supervised manner. Then, part-based features and object-based representations are integrated to produce the final classification. MAMC \cite{8} and Cross-X \cite{6} obtain specific part features directly through attention mechanisms with an end-to-end network. In addition, PMG \cite{42}, CA-PMG \cite{43} and DCL \cite{44} force models learn specific features from jigsaw patches. Zhang et al.\cite{86} leverages a small and clean meta-set to provide reliable prior knowledge for tackling noisy web images for webly-supervised FGVC. MetaIRNet\cite{87} combines generated images with original images to generate hybrid training images to improve the performance of one-shot FGVC. The above weakly supervised methods employ complicated structures and complex training strategies to extract key part features to reduce intra-class variances. However, there is no efficient way to handle inter-class differences in FGVC.
\subsection{Class Center}
The concept of class center is introduced to represent the whole class by center loss\cite{45, 77}, in which Euclidean distances between sample features and class centers are minimized to enhance the discriminative features in neural networks. Furthermore, Farzaneh et al.\cite{46} argued that not all elements in a class-center feature are relevant to discrimination, and further proposes sparse center loss. Specifically, the sparse center loss is calculated by multiplying the Euclidean distance in the center loss by the weights from an attention network. Li et al. also referred to the idea of class center and proposed single center loss (SCL)\cite{47}. SCL aggregates representations of natural samples around the center point and increases the distance from manipulated samples to the center point, making it greater than from natural samples by a margin. CSDL\cite{11} combines class centers and one-hot labels to generate soft labels. To measure the importance of samples in the same cluster, AdaMG\cite{78} calculates the distances between these samples and their corresponding class centers. Some few-shot methods represent the class as a whole with prototype \cite{88,89,90}, which is similar to the class center. However, considering intra-class variances, these methods with class center do not address the inter-class differences of FGVC, and do not fully explore for the ability of class centers. 
\subsection{Soft Labels}
One-hot labels are eligible for coarse-grained visual classification because of significant visual differences between coarse categories\cite{11}. However for FGVC, models with hard labels pay attention to irrelevant features (e.g., background) or sample-specific noise to achieve high prediction confidence from these "hard" labels. LS\cite{18} reduces prediction confidence by uniformly redistributing partial probabilities to nontarget classes to produce smoothed soft labels. Local distributional smoothness (LDS)\cite{48} proposes the local distributional smoothness of model outputs as a regularization term when inputs are perturbed. LR\cite{20} and GLR\cite{73} consider the rich inter-class correlations of FGVC, and optimize models with instance-level soft labels generated from a trained teacher network. However, the soft labels in these methods may be incorrect.

\begin{figure*}
	\centering
	\includegraphics[width=1.0\linewidth]{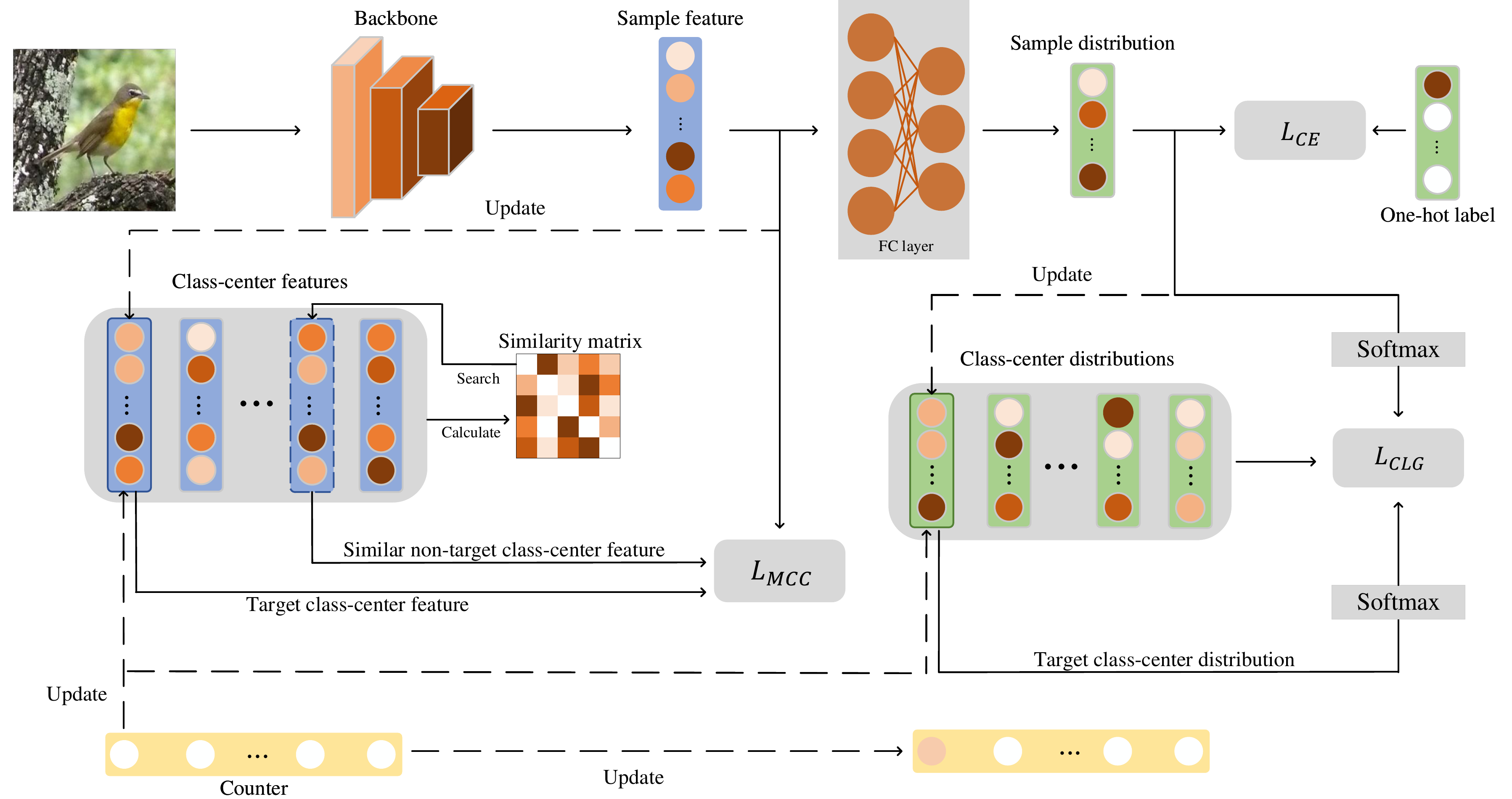}
	\caption{Overview of ECC. First, class-center features and class-center distributions are updated with sample features and sample distributions from the backbone with a counter. For intra-class variances, the MCC reduces the cosine distance between the sample feature and the target class-center feature. Moreover, for inter-class differences, the MCC enlarges the cosine distance between the sample feature and similar nontarget class-center feature which is determined by similarity matrix of class-center features. Moreover, class-center distributions are employed to generate soft labels with the softmax function. The KL divergence between soft labels and sample probability distributions is calculated as the CLG loss. The MCC and CLG are summed with hyperparameters $\lambda_1$ and $\lambda_2$ as ECC loss. Finally, the ECC loss is combined with the CE loss to supervise model.}\label{fig3}
\end{figure*}

\section{Exploration of Class Center}
In this section, we present the proposed ECC in detail. Our method handles problems of FGVC from the perspectives of features and labels. It includes: 1) an MCC which handles evident intra-class variances and subtle inter-class differences in the feature space, and 2) a CLG which addresses overfitting of models with reliable and reasonable soft labels. The framework of our ECC is shown in \textcolor{blue}{Fig.\ref{fig3}}. 

\subsection{Multiple Class-Center Constraint}
MCC optimizes feature distances between sample features and multiple class-center features, and aims to handle intra-class variances and inter-class differences. Given that the $i$-$th$ image belongs to the $y_{i}$-$th$ class, a basic neural network is utilized as a backbone to extract the $D$-dim feature $X_{i}\in\mathbb{R}^{D}$. 

First, class-center features $F_{y_{i}}\in\mathbb{R}^{D}$ are initialized and updated as the representation of the $y_{i}$-$th$ whole class. In most existing loss functions based on class-center \cite{45, 46}, class-center features are set as learnable parameters, and are updated backpropagation. Because of the small number of samples in each class of FGVC, it is difficult to stably learn robust class-center features. To address this issue, the MCC updates the class-center feature $F_{y_{i}}$ by averaging all the input sample features of the $y_{i}$-$th$ class in the training phase. Specifically, we first restore the sum of previously inputted sample features from the current class-center feature $F_{y_{i}}^{cur}$ of the $y_{i}$-$th$ class. For this purpose, a counter $C_{y_{i}}$ is maintained to record the number of sample features used for updating $F_{y_{i}}$. The sum of the previously input sample features of the $y_{i}$-$th$ class can be obtained by multiplying the current counter $C_{y_{i}}^{cur}$ by $F_{y_{i}}^{cur}$. Then, a new sample feature $X_{i}$ is added to the sum to update $F_{y_{i}}^{cur}$ as $F_{y_{i}}$. The updating of $F_{y_{i}}^{cur}$ and $C_{y_{i}}^{cur}$ is formulated as follows:
\begin{align}
	F_{y_{i}}&=\frac{1}{C_{y_{i}}^{cur}+1}\left(X_{i}+C_{y_{i}}^{cur}F_{y_{i}}^{cur}\right), \\
	C_{y_{i}}&=C_{y_{i}}^{cur}+1.
\end{align}
As the average of features of all samples in the $y_{i}$-$th$ class, $F_{y_{i}}$ does not need to be updated as learnable parameters in the training phase. Thus the updating is quite stable. The stability and effectiveness of our strategy are demonstrated in ablation studies.

Then we constrain the intra-class variances and inter-class differences with class-center features. To constrain intra-class variances, we reduce the feature distance between sample feature $X_i$ and the corresponding class-center feature $F_{y_{i}}$ directly. However, constraining only intra-class variances is not enough for FGVC. Subtle inter-class differences still cause confusion. Thus, for inter-class differences, we enlarge feature distance between $X_i$ and $F_{sim_{y_{i}}}$, which is the most similar class-center feature to $F_{y_{i}}$. $F_{sim_{y_{i}}}$ is obtained according to the similarity of class-center features. Specifically, we first construct a cosine similarity matrix $S \in \mathbb{R}^{N\times N}$, in which the similarity $s_{h,w}$ between the $h$-$th$ class and the $w$-$th$ class is calculated as
\begin{equation}
	\begin{split}
		s_{h, w}&=\cos \left(F_{h}, F_{w}\right) \\
		&=\frac{F_{h}^T \times F_{w}}
		{\|F_{h}\|_{2}\|F_{w}\|_{2}} \quad h,w \in\{0,1 \ldots, N\}.
	\end{split}
\end{equation}
$N$ denotes the number of classes. With the cosine similarity matrix $S$, the most similar class to the $y_{i}$-$th$ class is chosen by searching for the maximum value of the $y_{i}$-$th$ row in $S$:
\begin{equation}
	s_{{y_{i}}, sim_{y_{i}}}=\max \left(s_{y_{i}, 0}, s_{y_{i}, 1}, \ldots, s_{y_{i}, N}\right),
\end{equation}
where $sim_{y_{i}}$ is considered the index of the most similar class to the ${y_{i}}$-$th$ class.

In practice, the cosine distance $D_{cos}$ is utilized to represent the feature distance. Normally, $D_{cos}(X_i,F_{y_i})$ should be minimized in the trained phase, while $D_{cos} \left(X_i,F_{sim_{y_i}}\right)$ should be maximized. In order to integrate these two distances into a loss, cosine distance $D_{cos} \left(X_i,F_{sim_{y_i}}\right)$ is replaced with the cosine similarity $cos\left(X_i,F_{sim_{y_i}}\right)$. Moreover, $cos\left(X_i,F_{sim_{y_i}}\right)$ is multiplied by similarity $s_{y_i, sim_{y_i}}$ as a weight to adaptively handle confusing classes. For a mini-batch, the MCC loss function is expressed as
\begin{equation}
	\begin{split}
		L_{MCC}&\!=\!\sum_{k=1}^{M}\!\left(\!D_{cos}\!\left(\!X_{k},F_{y_{k}}\!\right)+s_{y_k,sim_{y_k}}\! cos\!\left(\!X_k,F_{sim_{y_k}}\!\right)\!\right)\!  \\
		&\!=\!\sum_{k=1}^{M}\!\left(\!1\!-\!cos\! \left(\!X_{k},F_{y_k}\!\right)+s_{y_k,sim_{y_k}} \!cos\!\left(\!X_{k},F_{sim_{y_{k}}}\!\right)\!\right)\! \\
		&\!=\!M\!+\!\sum_{k=1}^{M}\left(\!\frac{s_{y_{k},simy_{y_k}}\!X^T_{k}\!\!\times\! F_{sim_{y_k}}}{\|\!X_{k}\!\|_{2} \|\! F_{sim_{y_k}}\|_{2}} \!- \!\frac{X^T_{k}\!\times\! F_{y_{k}}}{\|\!X_{k}\!\|_{2} \|\! F_{y_k}\!\|_{2}} \!\right),
	\end{split}
\end{equation}
where $M$ denotes the number of samples in a mini-batch, $cos$ is the cosine similarity, and $k$ is the index of the sample in the mini-batch. 
 
\subsection{Class-Center Label Generation}
In this section, CLG is proposed to generate proper soft labels to alleviate overfitting and introduce relationships among classes. In our method, soft labels are generated from class-center distributions. Like class-center features, the class-center distribution $L_{y_i}\in\mathbb{R}^{N\times1}$ of the $y_i$-$th$ class is generated by averaging distributions of all input samples in the $y_i$-$th$ class during training phase. The updating strategy of the CLG can be expressed as follows:
\begin{equation}
	L_{y_i}=\frac{1}{C_{y_i}^{cur}+1}\left(f\left(X_{i}\right)+C_{y_i}^{cur}L_{y_i}^{cur}\right),
\end{equation}
where $f$ represents the last fully connected layer, which outputs a $N$-dim vector. Then, the sample probability distribution $P_{X_i}$ and the class-center probability distribution $Q_{y_i}$ are calculated with the $softmax$ activation function.  
\begin{align}
	P_{X_i}&=softmax(f(X_i))=[p_{i,1}, p_{i,2}, \ldots, p_{i,N}], \\
	Q_{y_i}&=softmax(L_{y_i})=[q_{y_i,1}, q_{y_i,2}, \ldots, q_{y_i,N}].
\end{align}

Consequently, the class-center probability distribution $Q_{y_i}$ is regarded as the soft label of $X_i$. The nontarget categories that are more similar to the target category have higher confidence in the soft labels. This nonuniform confidence is more reasonable and realistic.

For a mini-batch, KL divergence is computed between sample probability distributions and soft labels as the CLG loss:
\begin{equation}
	L_{CLG}=\sum_{k=1}^{M} KL\left(P_{X_k} \| Q_{y_{k}}\right)=\sum_{k=1}^{M} \sum_{n=1}^{N} p_{k,n} \log \frac{p_{k,n}}{q_{y_{k},n}}.
\end{equation}
$p_{k,n}$ and $q_{y_{k},n}$ denote the $n$-$th$ elements in $P_{X_k}$ and $Q_{y_k}$, respectively. With models supervised by soft labels of CLG, overfitting is effectively alleviated. Moreover, rich information among categories is considered to further improve the ability to model objects. 

\subsection{Exploration of Class Center}
Finally, the MCC and CLG are integrated as the ECC, and the two components are multiplied by the hyperparameters $\lambda_1$ and $\lambda_2$, respectively, to adjust the effects for model training. The entire ECC is formulated as,
\begin{equation}
	L_{ECC}=\lambda_1 L_{MCC}+\lambda_2 L_{CLG}.
\end{equation}

In addition, the CE loss $L_{CE}$ is combined with our ECC. The final loss function is expressed as follows:
\begin{equation}
	\begin{split}
		L_{final}&=L_{CE}+L_{ECC} \\
		&=-\frac{1}{M}\sum_{k=1}^{M}\sum_{n=1}^{N} l_{k,n} log(d_{k,n})+L_{ECC},
	\end{split}
\end{equation}
where $l_{k,n}$ denotes the $n$-$th$ element in the one-hot label of $X_k$, and $d_{k,n}$ is the $n$-$th$ element in the sample distribution $f(X_{k})$.

\section{Experiments}
\subsection{Implementation details}
Four widely-used fine-grained datasets are utilized in our experiments, including the AIR\cite{21}, CUB\cite{22}, CAR\cite{23} and NAB\cite{91} datasets. The details of datasets are shown in \textcolor{blue}{Table.\ref{table:table1}}. In addition, we conduct the experiments on a large-scale dataset iNaturalist 2018 (iNat2018)\cite{95}. The experimental results of iNat2018 are displayed and discussed in the supplementary material.

\begin{table}
	\caption{Statistics of datasets.}
	\centering
	\label{table:table1}
	\begin{tabular}{c|cccc}
		
		\hline
		\textbf{Dataset}  & \textbf{Object}   & \textbf{Classes} & \textbf{Train images} & \textbf{Test images} \\ \hline
		AIR\cite{21}      & Aircraft & 100     & 6667         & 3333        \\
		CUB\cite{22}      & Bird     & 200     & 5994         & 5794        \\
		CAR\cite{23}      & Car      & 196     & 8144         & 8041        \\
		NAB\cite{91}      & Bird     & 555     & 23,929       & 24,633        \\ \hline
	\end{tabular}
\end{table}

ResNet50\cite{24} is used as backbone in our experiments unless otherwise stated. For data augmentation, the images are resized to 600×600. Random cropping and center cropping are utilized to crop image to 448×448 during the training and test phases respectively. In addition, we apply random horizontal flipping in training. Stochastic Gradient Descent (SGD) is utilized with a momentum of 0.9. The initial learning rate is 0.01, which decays every 15 epochs at a decay rate of 0.1. The initial learning rate is multiplied by 0.1 for the pretrained backbone on the CUB dataset. The batch size and epochs are set as 32 and 50, respectively. Class-center features and class-center distributions are randomly initialized before training. For the center loss, PC loss \cite{15} and LS, we choose the optimal hyperparameters among the settings from original papers and our experimental best values. Any extra annotations or extra training data are not used and all backbone models are pretrained on the ImageNet dataset. 

In ablation studies, same training hyperparameters (including batch size, learning rate and so on) are utilized. And Resnet50 is used as the backbone for all ablation experiments. 

\subsection{Integration with existing FGVC methods and different backbones}
First, to verify the effectiveness of our method, we test it on different backbones including InceptionV3\cite{18}, ResNet50, ResNet101\cite{24} and DenseNet121 \cite{59}. According to \textcolor{blue}{Table.\ref{table:table5}}, our ECC brings satisfactory improvements on four datasets (0.8\%$\sim$1.9\% on AIR, 0.6\%$\sim$3.4\% on CUB, 1.0\%$\sim$1.6\% on CAR and 1.2\%$\sim$3\% on NAB). Compared with the improvements on InceptionV3 (0.8\% on AIR, 0.6\% on CUB, 1.2\% on CAR, 1.2\% on NAB), ResNet50, DenseNet121 and ResNet101 have better feature extraction capabilities for constructing better class centers. Thus, there are more improvements on those models (average boost of 1.83\% on AIR, 2.8\% on CUB, 1.4\% on CAR and 2.9\% on NAB). 

We also show the results of integration with existing FGVC methods, including DCL\cite{44}, MGE-CNN\cite{39}, WSDAN\cite{4}, Swin transformer \cite{50} and CAL\cite{49} in \textcolor{blue}{Table.\ref{table:table6}}. Compared with baselines, integrated methods obtain obvious and consistent improvements on these four datasets.

\begin{table}
	\caption{Integration with different backbones.The best results are shown in bold.}
	\centering
	\label{table:table5}
	\begin{tabular}{c|c|cccc}
		\hline
		\textbf{Model} &\textbf{Loss}  &\textbf{AIR} &\textbf{CUB}  & \textbf{CAR}  & \textbf{NAB}\\
		\hline
		InceptionV3		   &CE loss &90.7 &83.9  &92.8 &83.3\\
		InceptionV3-ECC    &ECC loss &\textbf{91.5}	&\textbf{84.5}	&\textbf{94.0}	&\textbf{84.5}\\
		\hline
		ResNet50	    &CE loss	&91.1	&84.7	&93.1 &83.4\\
		ResNet50-ECC	&ECC loss	&\textbf{93.0} &\textbf{87.3}	&\textbf{94.7} &\textbf{85.5}\\
		\hline
		DenseNet121     &CE loss	&91.6	&84.6	&93.0 &83.8\\
		DenseNet121-ECC &ECC loss	&\textbf{93.4}	&\textbf{88.0}	&\textbf{94.6} &\textbf{86.2}\\
		\hline
		ResNet101	  &CE loss	&91.5	&85.6	&93.7 &83.5\\
		ResNet101-ECC &ECC loss	&\textbf{93.3}	&\textbf{88.0}	&\textbf{94.7 } &\textbf{86.5}\\
		\hline
	\end{tabular}
\end{table} 

\begin{table}
	\caption{Integration with different FGVC methods.The best results are shown in bold.}
	\centering
	\label{table:table6}
	\begin{tabular}{c|c|cccc}
		\hline
		\textbf{Method} &\textbf{Backbone} & \textbf{AIR} &\textbf{CUB} & \textbf{CAR} &\textbf{NAB} \\
		\hline
		DCL\cite{44}          &ResNet50    &93.0	        &87.8	        &94.5              &86.0\\
		DCL-ECC               &ResNet50    &\textbf{93.7}	&\textbf{88.8}  &\textbf{95.0}     &\textbf{87.2}\\
		\hline
		MGE-CNN\cite{39}      &ResNet50    & -              &88.5           &93.9              &86.7 \\
		MGE-CNN-ECC           &ResNet50    &\textbf{93.8}	&\textbf{88.8}	&\textbf{94.8}     &\textbf{86.9} \\
		\hline
		WSDAN\cite{4}         &InceptionV3 &93.0	        &89.4	        &94.5              &87.9\\
		WSDAN-ECC             &InceptionV3 &\textbf{94.0}	&\textbf{89.7}	&\textbf{94.8}     &\textbf{88.7} \\
		\hline
		Swin\cite{50}         &Swin-base   &92.2	        &91.0	        &94.5              &90.7\\
		Swin-ECC              &Swin-base   &\textbf{92.8}	&\textbf{92.3}	&\textbf{94.7}     &\textbf{91.4} \\
		\hline
		CAL\cite{49}          &ResNet101   &94.2	        &90.6	        &95.5              &91.0\\
		CAL-ECC	              &ResNet101   &\textbf{95.2}	&\textbf{91.0}	&\textbf{95.9}     &\textbf{91.3} \\
		\hline
	\end{tabular}
\end{table}

\begin{table*}
	\caption{Comparison with different loss functions.The best results are shown in bold.}
	\centering
	\label{table:table7}
	\begin{tabular}{c|c|cccccccc}
		\hline
		\textbf{Backbone} &\textbf{Dataset} &\textbf{Baseline} &\textbf{Ct loss} &\textbf{SC loss} &\textbf{PC loss} &\textbf{LS} &\textbf{Ctt loss} &\textbf{Tlt loss} &\textbf{Ours}\\
		\hline
		\multirow{4}*{InceptionV3} 		   
					&AIR &90.7 &90.8 &90.5 &90.7 &90.7 &91.3 &91.3 &\textbf{91.5} \\
		            &CUB &83.9 &84.2 &82.1 &84.7 &83.6 &\textbf{85.1} &\textbf{85.1} &84.5 \\
					&CAR &92.8 &92.6 &92.9 &93.0 &92.8 &93.2 &93.5 &\textbf{94.0} \\
					&NAB &83.3 &83.4 &83.4 &83.9 &83.8 &83.5 &84.1 &\textbf{84.5} \\
		\hline
		\multirow{4}*{ResNet50}			
					&AIR &91.1 &91.5 &91.5 &91.6 &92.1 &91.8 &92.1 &\textbf{93.0} \\
		            &CUB &84.7 &84.9 &86.2 &85.5 &85.5 &86.3 &86.5 &\textbf{87.3} \\
					&CAR &93.1 &93.2 &93.0 &93.8 &94.1 &93.2 &93.9 &\textbf{94.7} \\
					&NAB &83.4 &83.4 &84.5 &84.2 &85.3 &83.6 &84.3 &\textbf{85.5} \\
		\hline
		\multirow{4}*{DenseNet121}			
					&AIR &91.6 &91.8 &91.6 &91.6 &92.2 &91.8 &92.5 &\textbf{93.4} \\
		            &CUB &84.6 &84.5 &86.4 &86.0 &85.3 &86.5 &86.3 &\textbf{88.0} \\
					&CAR &93.0 &92.8 &93.0 &93.3 &93.7 &93.4 &94.3 &\textbf{94.6} \\
					&NAB &83.8 &84.0 &85.4 &84.7 &85.3 &84.3 &85.0 &\textbf{86.2} \\
		\hline
		\multirow{4}*{ResNet101}			
					&AIR &91.5 &91.9 &91.8 &91.9 &92.3 &92.1 &92.5 &\textbf{93.3} \\
	                &CUB &85.6 &85.7 &85.8 &86.4 &86.5 &87.0 &87.0 &\textbf{88.0} \\
					&CAR &93.7 &93.6 &93.8 &94.0 &94.2 &93.4 &94.3 &\textbf{94.7} \\
					&NAB &83.5 &84.6 &85.2 &85.5 &86.4 &84.8 &85.9 &\textbf{86.5} \\
		\hline
	\end{tabular}
\end{table*}

\subsection{Comparison with different loss functions}
In this section, we compare our approach with different loss functions, including Center loss (Ct loss)\cite{45}, Single Center loss (SC loss)\cite{47}, Pairwise Confusion loss (PC loss)\cite{15}, Label Smoothing (LS)\cite{18},  Contrastive loss (Ctt loss)\cite{74} and Triplet loss (Tlt loss)\cite{75} on four datasets. For fairness, experiments are conducted with different hyperparameters (including recommended hyperparameters from original papers and our experimental best values) on ResNet50 to choose the optimal results for comparison with our methods. Moreover, ResNet50 is replaced with different backbones (Inceptionv3, DenseNet121, ResNet101) to further demonstrate the superiority of our method. Results are shown in \textcolor{blue}{Table.\ref{table:table7}}. CE loss is regarded as baseline, which have achieved acceptable results. Existing methods based on class center (Ct loss, SC loss) bring few improvement, due to the unstable updating strategy of class centers and the lack of constraints for inter-class differences. Contrastive loss and Triplet loss (Ctt loss and Tlt loss) achieve better performances than methods based on class center. Compared with the above methods, our method effectively constrains intra-class variances and inter-class differences and achieves optimal overall performance.
 
An exception is InceptionV3 on CUB dataset, where Ctt loss and Tlt loss are superior to our MCC (85.1\% vs 84.5\%). In fact, in our method, the quality of the class-center features and class-center distributions depends on the quality of the sample features and sample distributions extracted from the backbones. Compared with other backbones (ResNet50, ResNet101 and DenseNet121), the features and distributions from InceptionV3 are not good enough, which leads to limited improvement. In practice, compared with InceptionV3, ResNet and DenseNet have more extensive applications, and most FGVC methods use ResNet and DenseNet as backbones. With these backbones, our method has more obvious advantages. Therefore, our MCC loss has greater application value in FGVC.

\subsection{Comparison with SoTA methods}

\begin{table}
	\caption{Comparison with SoTA methods.The best results are shown in bold.} 
	\centering
	\label{table:table8}
	\begin{tabular}{c|cccc}
		\hline
		\textbf{Method} & \textbf{AIR} & \textbf{CUB} & \textbf{CAR} & \textbf{NAB}\\
		\hline
		B-CNN\cite{51} &86.9 &84.0 &90.6 &-\\
		MA-CNN\cite{2} &89.9 &86.5 &92.8 &-\\
		M2DRL\cite{93} & -  &87.2 &93.3 &-\\
		NTS-Net\cite{94} & 91.4  &87.5 &93.9 &-\\
		Cross-X\cite{6}&92.6 &87.7 &94.6 &86.2\\
		MGE-CNN\cite{39} & -  &88.5  &93.9 &86.7\\
		ELP\cite{57}  &92.7 &88.8 &94.2 &-\\
		DCL\cite{44}  &93.0 &87.8 &94.5 &86.0\\
		WSDAN\cite{4} &93.0 &89.4 &94.5 &87.9\\
		SFFF\cite{76}   &93.1 &85.4 &94.4 &-\\
		API-Net\cite{70} &93.4 &88.6 &94.9 &86.2\\
		PMG\cite{42}     &93.4 &89.6 &95.1 &-\\
		CDSL-DCL\cite{11}&93.5 &88.6 &94.9 &-\\
		CAL\cite{49}     &94.2 &90.6 &95.5 &91.0\\
		SIA-Net\cite{82} &94.3 &90.7 &95.5 &-\\
		ALIGN\cite{61} &- &- &\textbf{96.1} &-\\
		\hline
		ViT\cite{25}    &- &90.3 &93.7 &89.9\\
		AA-Trans\cite{85}& -   &91.4 & - &90.2\\
		TransFG\cite{10}& -   &91.7 &94.8 &90.8\\
		Swin\cite{50}   &92.2 &91.0 &94.5 &90.7\\
		CAMF\cite{69}   &93.3 &91.2 &95.3 &-\\
		SIM-Trans\cite{92} & - &91.8 & - &-\\
		Dual-TR\cite{77}& -   &92.0 & -   &91.3\\	
		\hline
		DCL-ECC        &93.7 &88.8	&95.0 &87.2\\
		MGE-CNN-ECC    &93.8 &88.8 &94.8 &86.9\\
		WSDAN-ECC      &94.0 &89.7 &94.8 &88.7\\		
		CAL-ECC	       &\textbf{95.2} &91.0 &95.9 &91.3\\
		Swin-ECC       &92.8 &\textbf{92.3} &94.7 &\textbf{91.4}\\
		\hline
	\end{tabular}
\end{table}

In this section, integrated models are compared with existing state-of-the-art (SoTA) approaches. Extensive experiments are conducted to verify the effectiveness of our ECC. \textcolor{blue}{Table.\ref{table:table8}} shows the performances. Compared with other methods, our ECC (Swin-ECC on the CUB dataset, CAL-ECC on the AIR and CAR datasets, and Swin-ECC on the NAB dataset) achieves excellent performances, with SoTA on the CUB and AIR datasets. Although ALIGN outperforms our method on CAR dataset, ALIGN is pretrained on a large-scale noisy image-text dataset (LSNITD)\cite{61}, which includes 1.8B image-text pairs. ImageNet-1k and ImageNet-21k are utilized in our integrated models. The size of LSNITD is 120 times larger than  ImageNet-21k. 

Moreover, we made an interesting observation. Transformer-based methods achieve better results than CNNs on the CUB and NAB datasets, but CNNs achieve better performances on the AIR and CAR datasets. We argue that transformer-based methods are naturally not subject to the local inductive bias of CNNs. Thus, these methods have the ability to model global dependency, which has advantages over CNN-based methods in terms of classifying nonstructural rigid objects (e.g., birds). However, transformer-based methods destroy the structural information of rigid structural objects including cars and aircraft, by preprocessing an image into a sequence of flattened patches \cite{54}. This process leads to inferior performance on the AIR and CAR datasets.

\subsection{Ablation studies}
In this section, ablation studies are conducted for different components of our ECC. First, hyperparameters of the MCC and CLG are investigated via extensive experiments. Then, each component in our approach is explored. In addition, we discuss different updating strategies of class centers in detail.

\subsubsection{Hyperparameter Selection}

\begin{figure}
	\centering
	\includegraphics[width=1.0\linewidth]{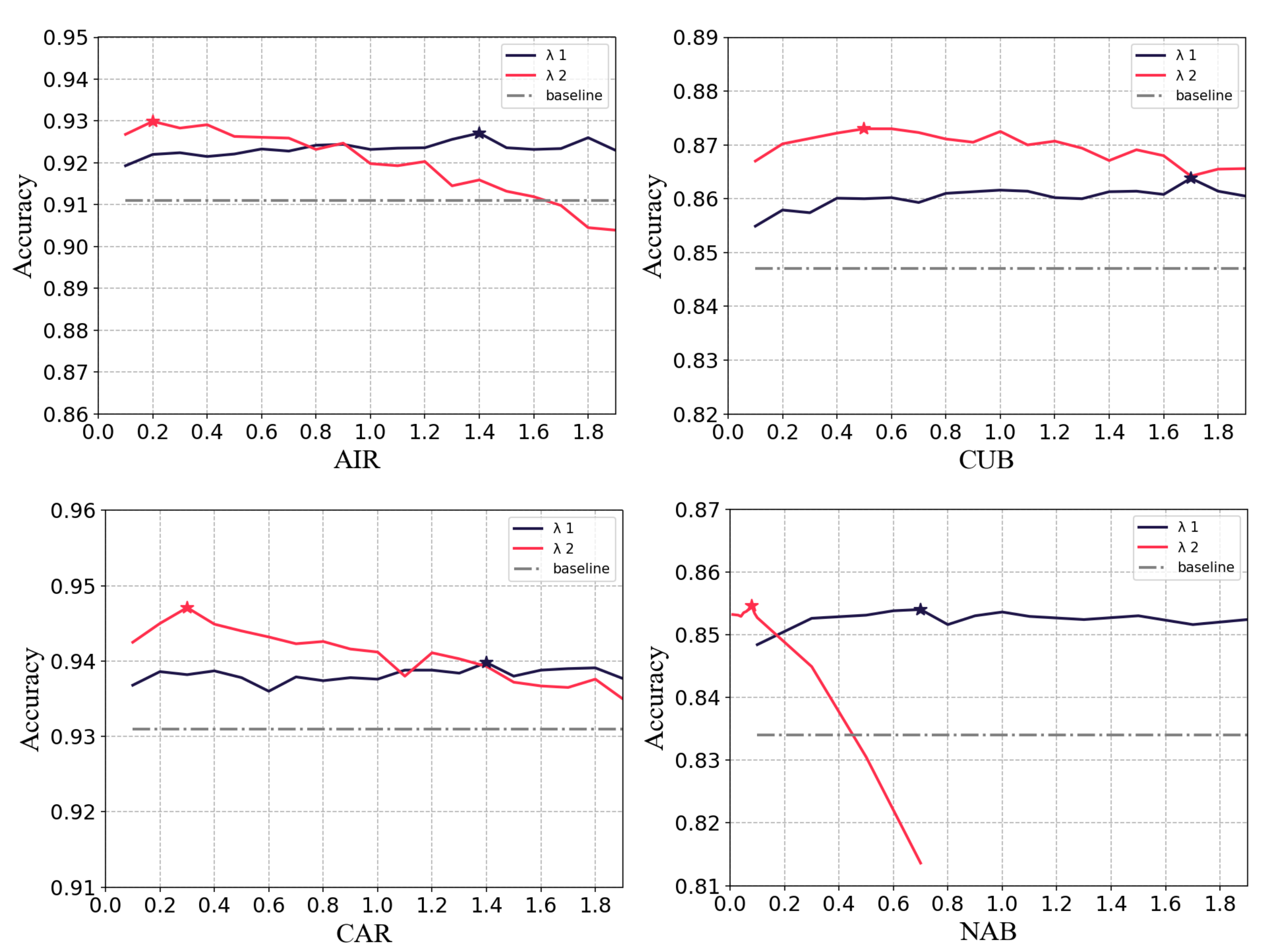}
	\caption{The performances of different $\lambda_1$ for the MCC and $\lambda_2$ for the CLG. Baselines are represented by dashed grey lines. The blue curves correspond to the changes in the MCC weight $\lambda_1$. The red curves correspond to the changes in the CLG weight $\lambda_2$.}\label{fig4}
\end{figure}

To determine the proper weights for proposed components, we test the performances of different values on all datasets. The best results with MCC occur at $\lambda_1$=1.4 on the AIR, $\lambda_1$=1.7 on CUB, $\lambda_1$=1.4 on CAR and $\lambda_1$=0.7 on NAB. With chosen $\lambda_1$, $\lambda_2$ are chosen for CLG as 0.2 on AIR, 0.6 on CUB, 0.3 on CAR and 0.08 on NAB. 

The changes in accuracy are displayed in \textcolor{blue}{Fig.\ref{fig4}}. When $\lambda_1$ increases, the accuracy changes less than $1\%$ for all datasets ($0.8\%$ for AIR, $0.9\%$ for CUB, $0.4\%$ for CAR, and $0.6\%$ for NAB). Regardless of what value is used for $\lambda_1$ from 0.1 to 2.0, our MCC is always better than that of the baselines (represented by dashed grey lines in \textcolor{blue}{Fig.\ref{fig4}}), which may indicate that the MCC is not sensitive to $\lambda_1$ and that better parameters can lead to better results. However, there is an obvious change when $\lambda_2$ increases. This change indicates that the CLG is more sensitive to weight than the MCC. In the early stage of training, the class-center features and class-center distributions are both unreliable, which results in incorrect supervision information for the model. However, in the MCC, it is also reasonable to expand the distance between sample features and unreliable similar class-center features. Thus, the impact of the weight of the MCC is relatively small. Unreliable class-center distributions have a greater effect on the CLG than on the MCC. Therefore, the weight of the CLG should be smaller. In addition, the size of the NAB dataset is three to four times greater than that of other datasets. Therefore, there are more erroneous predictions at the early stage of training than with other datasets, and the early class-center distributions are more unreliable. A larger weight leads to a more serious negative impact on the model. 

\subsubsection{Contribution of Components}
\textcolor{blue}{Table.\ref{table:table9}} shows the performance of the proposed components and their combinations. First, the performance of each separate component (MCC and CLG) is demonstrated. Compared with the baseline (CE loss), our method shows obvious advances, namely, 1.6\% on AIR, 1.7\%$\sim$2.4\% on CUB, 0.9\%$\sim$1.5\% on CAR and 1.0\%$\sim$2.0\% on NAB. Finally, our ECC obtains excellent results with a combination of two components (93.0\% on AIR, 87.3\% on CUB, 94.7\% on CAR and 85.5\% on NAB). The improvements verify the effectiveness of all the components. The MCC, CLG and their combination all boost the performances significantly on all datasets.

\begin{table}
	\caption{Contribution of proposed components and their combinations.The best results are shown in bold.}
	\centering
	\label{table:table9}
	\begin{tabular}{c|c|cccc}
		\hline
		\textbf{Component} & \textbf{Backbone} & \textbf{AIR} & \textbf{CUB} & \textbf{CAR} & \textbf{NAB}\\
		\hline
		Baseline&ResNet50 &91.1 &84.7 &93.1 &83.4\\
		MCC	&ResNet50 &92.7 &86.4 &94.0 &85.4\\
		CLG	&ResNet50 &92.7	&87.1 &94.6 &84.4\\
		ECC	&ResNet50 &\textbf{93.0} &\textbf{87.3} &\textbf{94.7} &\textbf{85.5}\\
		\hline
	\end{tabular}
\end{table}

\begin{figure}
	\centering
	\includegraphics[width=1.0\linewidth]{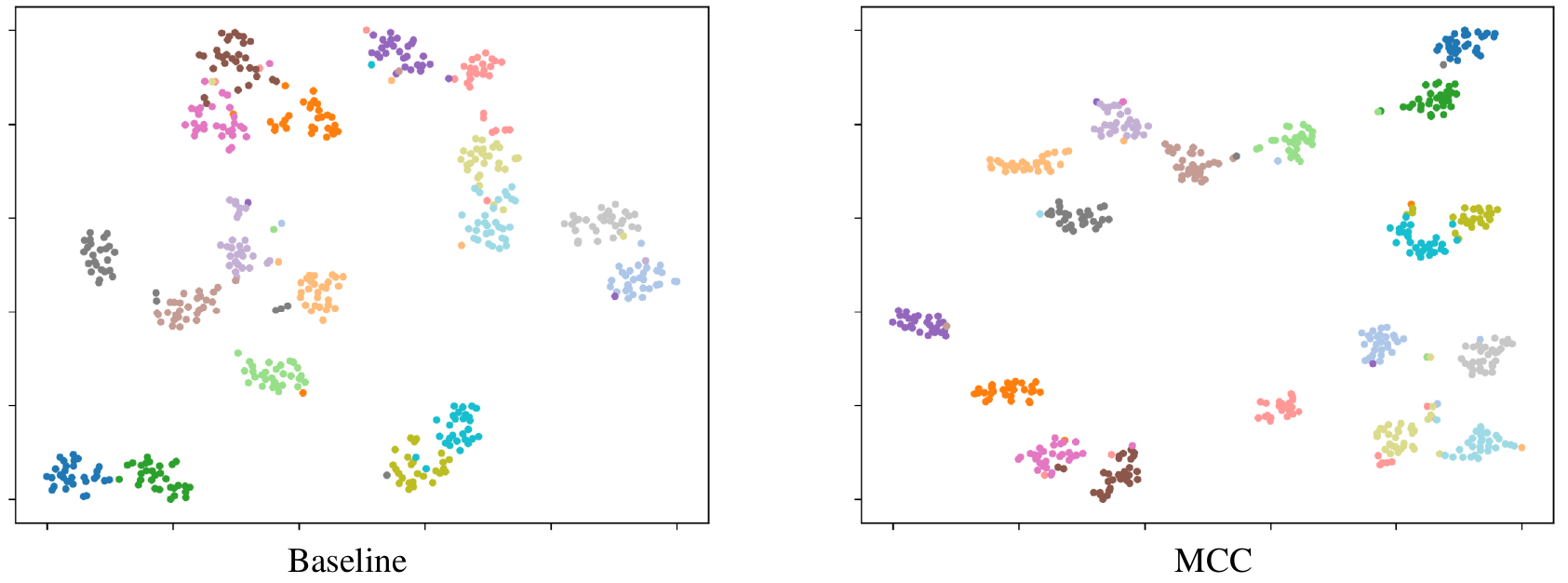}
	\caption{The visualizations of t-SNE on 18 species of visually similar warblers from the CUB dataset. The left image represents the result of the CE loss. The right image is the result of the MCC. Points with the same colour belong to one class.}\label{fig5}
\end{figure}

\begin{figure*}
	\centering
	\includegraphics[width=1.0\linewidth]{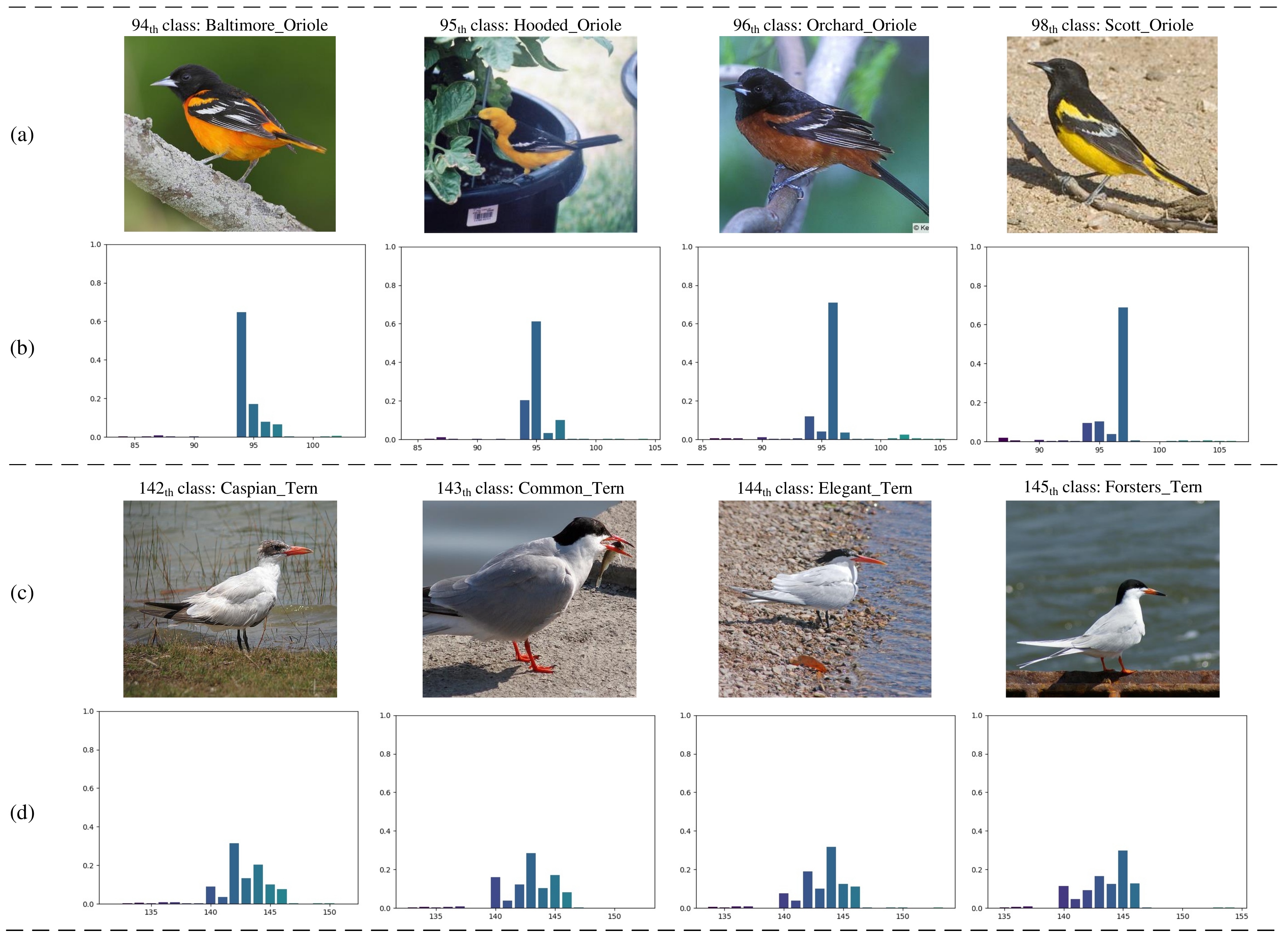}
	\caption{Soft labels of some similar classes in the CUB dataset. (a) and (c) denote original images belonging to similar classes. (b) and (d) are the corresponding soft labels of (a) and (c) in the CLG. For simplicity, only 20 classes among the target classes are included in each figure.}\label{fig6}
\end{figure*}

\textcolor{blue}{Fig.\ref{fig5}} shows the results of t-SNE \cite{55} on the CUB dataset. The left figure and right figure show the results before and after adding our MCC, respectively. In the left figure, many points belonging to the same class (with the same colour) are scattered and mixed with points from other classes. The results of adding the MCC are displayed in the right figure and compared with those in the left figure. These scattered points obviously converge after adding the MCC. Moreover, there are no clear boundaries between some extremely confusing classes with CE loss, while clear boundaries appear constrained by the MCC. This finding suggests that our MCC component can compress intra-class variances and expand inter-class differences simultaneously. 

The soft labels are displayed in \textcolor{blue}{Fig.\ref{fig6}}. The images in row (a) all belong to orioles, and there are subtle differences. Therefore, the four classes are visually similar, and our soft labels also reflect the similarities between classes in row (b). Rows (c) and (d) also support the reasonableness of our soft labels.

\subsubsection{Discussion of the Strategy of Updating the Class Center}
\begin{table}
	\caption{Different the strategies of updating the class center.The best results are shown in bold.}
	\centering
	\label{table:table10}
	\begin{tabular}{c|c|cccc}
		\hline
		\textbf{Method} &\textbf{Backbone} &\textbf{AIR} &\textbf{CUB}  &\textbf{CAR} &\textbf{NAB} \\
		\hline
		Baseline                   &ResNet50 &91.1 &84.7 &93.1 &83.4 \\
		Center                     &ResNet50 &91.5 &84.9 &93.2 &83.4 \\
		Contrastive center         &ResNet50 &91.3 &85.9 &93.5 &83.4 \\
		Deep attentive center      &ResNet50 &91.7 &86.0 &93.5 &84.3 \\
		Ours     	               &ResNet50 &\textbf{93.0} &\textbf{87.3} &\textbf{94.7} &\textbf{85.5}\\
		\hline
	\end{tabular}
\end{table}
\begin{figure}
	\centering
	\includegraphics[width=1.0\linewidth]{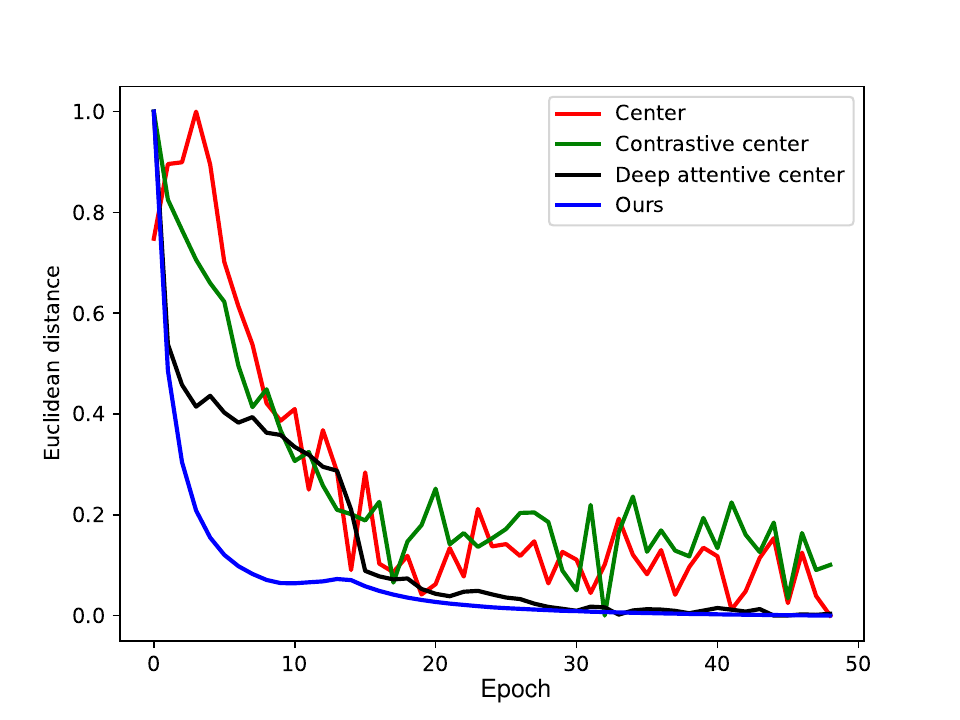}
	\caption{Stability comparison of several methods based on class center and our strategy on the CUB dataset. We utilize the Euclidean distance of class-center features between each epoch and its next epoch to measure the stability of the update of the class center.}\label{fig7}
\end{figure}
To verify the advantage of our proposed strategy of updating class center, we compare different loss functions based on the class center, including center loss (Ct loss), contrastive center loss (CttCt loss)\cite{71} and deep attentive center loss (DACt loss)\cite{46}. The Euclidean distances of class-center features between each epoch and its next epoch, are utilized to represent changes of class-center features. \textcolor{blue}{Fig.\ref{fig7}} shows the results on the CUB dataset. There are violent shakes during the training phase with the strategy of center loss. Contrastive center loss further handles inter-class differences, but does not optimize the update strategy. Thus, violent shaking still occurs. The deep attentive center loss utilizes attention to reduce the impact of useless elements in the class center. Although deep attentive center loss further stabilizes the update of class center, it does not actually solve the problem. Compared with the above methods, our strategy always maintains a smooth curve throughout the training phase. This finding indicates that our strategy facilitates the stable updating of the class center. Furthermore, we compare the performances of different updating strategies for class center in \textcolor{blue}{Table.\ref{table:table10}}. Consistent superior performance also demonstrates the advantage of our strategy.

\begin{table}
	\caption{Comparison of computational complexity before and after using our methods.}
	\centering
	\label{table:table11}
	\begin{tabular}{c|cccc}
		\hline
		\textbf{Model} &\textbf{AIR} &\textbf{CUB}  & \textbf{CAR}  & \textbf{NAB}\\
		\hline
		InceptionV3		   &13.21G &13.21G  &13.21G &13.21G\\
		InceptionV3-ECC    &13.28G	&13.46G	&13.45G	&15.11G\\
		\hline
		ResNet50	    &16.44G	&16.44G	&16.44G &16.44G\\
		ResNet50-ECC	&16.50G &16.69G	&16.68G &18.34G\\
		\hline
		DenseNet121     &11.46G	&11.46G	&11.46G &11.46G\\
		DenseNet121-ECC &11.49G	&11.58G &11.58G &12.41G\\
		\hline
		ResNet101	  &31.33G	&31.33G	&31.33G &31.33G\\
		ResNet101-ECC &31.39G	&31.39G	&31.39G &31.39G\\
		\hline
		Average extra FLOPs &0.06G	&0.22G	&0.24G &1.66G\\
		\hline
	\end{tabular}
\end{table} 
\begin{figure*}
	\centering
	\includegraphics[width=0.8\linewidth]{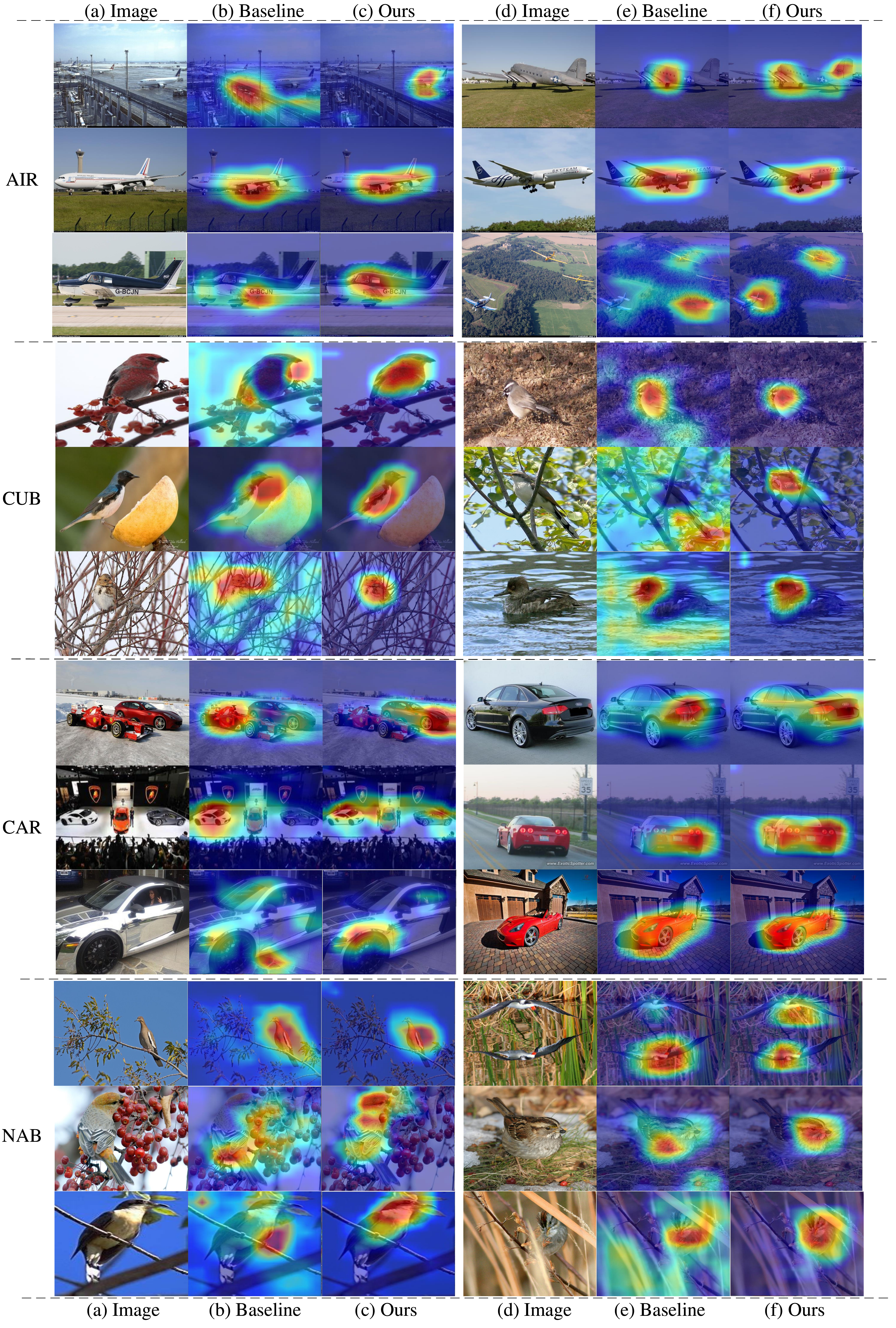}
	\caption{The visualization of Grad-CAM \cite{56} on the AIR, CUB, CAR and NAB datasets. (a) and (d) are original images. (b) and (e) show heatmaps of the baseline (CE loss). (c) and (f) are heatmaps of our ECC.}\label{fig8}
\end{figure*}

\begin{figure}
	\centering
	\includegraphics[width=1.0\linewidth]{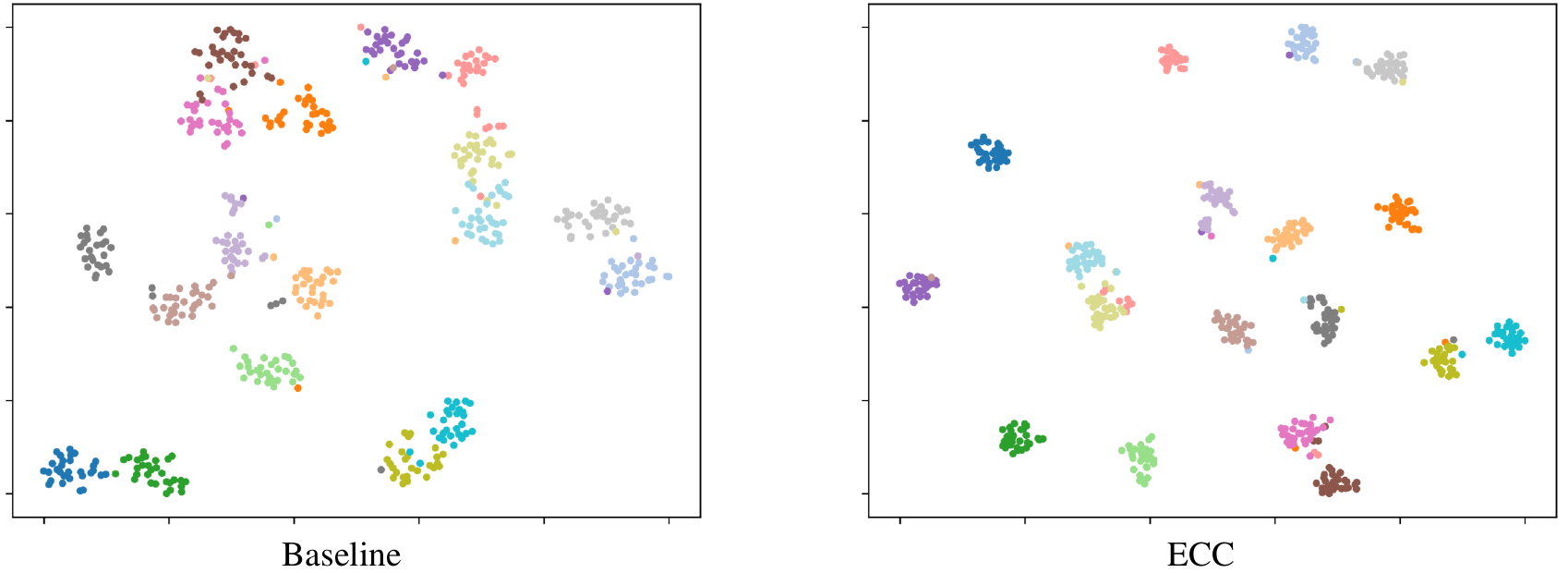}
	\caption{The t-SNE visualizations of 18 species of visually similar warblers from the CUB dataset. The first row and the second row show the results of the baseline and our ECC, respectively. Points with the same colour belong to one class.}\label{fig9}
\end{figure}

\subsection{Discussion of Time Complexity and Computational Complexity}

In this section, we consider the complexity of our ECC. First, the time complexity is discussed. To update the class center feature, the MCC needs to index the class center feature with the corresponding label. The class-center feature is updated according to Eq. (1). The time complexity of updating the class center feature is $O(1)$. To handle inter-class differences, the MCC needs to calculate the similarity between each class according to Eq. (3), whose time complexity is $O(N^2)$, where $N$ is the number of classes. Then, the MCC algorithm searches the maximum in a row of the similarity matrix, with a time complexity of $O(N)$. Finally, the MCC loss is calculated according to Eq. (5) with a time complexity of $O(1)$. In summary, the time complexity of MCC is $O(1)+O(N^2)+O(N)+O(1) \approx O(N^2)$. Furthermore, the CLG indexes the class center label with the corresponding label to update the class center label according to Eq. (6) and calculates the loss according to Eq. (9). The time complexity of the CLG is $O(1)$. The overall time complexity is determined by the complexity of the MCC and the CLG. Therefore, the overall time complexity of our method is $O(N^2)+O(1) \approx O(N^2)$.

Second, we discuss the computational complexity and calculate the FLOPs of the MCC and CLG. Actually, the computational complexity of the CLG is less than that of 0.1 G FLOPs, which is negligible considering that of the FLOPs of the MCC. Therefore, we regard the FLOPs of the MCC as the overall FLOPs. The comparison of the computational complexity before and after using our method is shown in \textcolor{blue}{Table.\ref{table:table11}}. Our method requires only a few FLOPs (average 0.06 G to 1.66 G on all datasets). Compared with the complexity of the backbone network, this computational complexity is insignificant. Moreover, the costs exist in the training phase only and are not incurred in the practical test phase. Thus, our method is very practical and can result in significant improvement with negligible costs.

\subsection{Visualization}

First, we visualize heatmaps with Grad-CAM\cite{56} images, which are shown in \textcolor{blue}{Fig.\ref{fig8}}. Our ECC guides the model to learn discriminative features and alleviates model overfitting. It is evident that the model no longer pays attention to background information, especially in the CUB and NAB datasets, which usually contain complex backgrounds. Moreover, in the first and last heatmaps of the AIR dataset, the results of CE loss incorrectly focus on complex background information, while our method correctly focuses on the objects. Similarly, our method achieves better results on the CAR dataset. \textcolor{blue}{Fig.\ref{fig9}} displays the t-SNE results of the baseline and our ECC loss on 18 species of visually similar warblers from the CUB dataset. The left image and right image represent the results of CE loss and our ECC loss, respectively. There are more evident margins between different classes in the results of ECC than in those of CE loss. Compression within a class is also observed in the t-SNE results. These results indicate the effectiveness of our method.

\section{Conclusion}
In this paper, we propose a simple but effective method named ECC to improve the feature extraction capability of the model. ECC explores the role of class centers from the perspectives of features and labels with two components: an MCC and a CLG. From the feature perspective, the MCC reduces intra-class variances by reducing the cosine distance between sample features and target class-center features. Moreover, the MCC decreases the cosine similarity between sample features and the most similar nontarget class-center features to increase intra-class differences. Furthermore, from the label perspective, the CLG converts the class-center distribution of each class as a soft label to supervise the model to alleviate overfitting. Our soft labels are reliable and introduce correlations between categories. Finally, ECC loss and CE loss are combined to optimize the model. Extensive experiments and visualizations demonstrate the effectiveness of our method.

\section*{Acknowledgements}

The work was jointly supported by the National Science and Technology Major Project under grant No. 2022ZD0117103, the National Natural Science Foundations of China under grant No. 62272364, the provincial Key Research and Development Program of Shaanxi under grant No. 2024GH-ZDXM-47, the Teaching Reform Project of Shaanxi Higher Continuing Education under Grant No. 21XJZ004, the Innovation Fund of Xidian University, High-performance Computing Platform of XiDian University, Natural Science Basic Research Program of Shaanxi (Program No. 2024JC-YBQN-0639).

\ifCLASSOPTIONcaptionsoff
  \newpage
\fi

\bibliographystyle{IEEEtran}
\bibliography{cas-refs}

\begin{thebibliography}{10}
\providecommand{\url}[1]{#1}
\csname url@samestyle\endcsname
\providecommand{\newblock}{\relax}
\providecommand{\bibinfo}[2]{#2}
\providecommand{\BIBentrySTDinterwordspacing}{\spaceskip=0pt\relax}
\providecommand{\BIBentryALTinterwordstretchfactor}{4}
\providecommand{\BIBentryALTinterwordspacing}{\spaceskip=\fontdimen2\font plus
\BIBentryALTinterwordstretchfactor\fontdimen3\font minus
  \fontdimen4\font\relax}
\providecommand{\BIBforeignlanguage}[2]{{%
\expandafter\ifx\csname l@#1\endcsname\relax
\typeout{** WARNING: IEEEtran.bst: No hyphenation pattern has been}%
\typeout{** loaded for the language `#1'. Using the pattern for}%
\typeout{** the default language instead.}%
\else
\language=\csname l@#1\endcsname
\fi
#2}}
\providecommand{\BIBdecl}{\relax}
\BIBdecl

\bibitem{1}
O.~Russakovsky, J.~Deng, H.~Su, J.~Krause, S.~Satheesh, S.~Ma, Z.~Huang,
  A.~Karpathy, A.~Khosla, M.~Bernstein, A.~C. Berg, and L.~Fei-Fei, ``{ImageNet
  Large Scale Visual Recognition Challenge},'' \emph{International Journal of
  Computer Vision}, vol. 115, no.~3, pp. 211--252, 2015.

\bibitem{4}
\BIBentryALTinterwordspacing
T.~Hu, H.~Qi, Q.~Huang, and Y.~Lu, ``{See Better Before Looking Closer: Weakly
  Supervised Data Augmentation Network for Fine-Grained Visual
  Classification},'' 2019. [Online]. Available:
  \url{http://arxiv.org/abs/1901.09891}
\BIBentrySTDinterwordspacing

\bibitem{7}
Z.~Huang and Y.~Li, ``{Interpretable and accurate fine-grained recognition via
  region grouping},'' in \emph{Proceedings of the IEEE Computer Society
  Conference on Computer Vision and Pattern Recognition}, 2020, pp. 8659--8669.

\bibitem{39}
L.~Zhang, S.~Huang, W.~Liu, and D.~Tao, ``{Learning a mixture of
  granularity-specific experts for fine-grained categorization},'' in
  \emph{Proceedings of the IEEE International Conference on Computer Vision},
  vol. 2019-Octob, 2019, pp. 8330--8339.

\bibitem{76}
M.~Wang, P.~Zhao, X.~Lu, F.~Min, and X.~Wang, ``Fine-grained visual
  categorization: A spatial--frequency feature fusion perspective,'' \emph{IEEE
  Transactions on Circuits and Systems for Video Technology}, 2022.

\bibitem{2}
H.~Zheng, J.~Fu, T.~Mei, and J.~Luo, ``{Learning Multi-attention Convolutional
  Neural Network for Fine-Grained Image Recognition},'' in \emph{Proceedings of
  the IEEE International Conference on Computer Vision}, vol. 2017-Octob, 2017,
  pp. 5219--5227.

\bibitem{3}
\BIBentryALTinterwordspacing
J.~Zhang, R.~Zhang, Y.~Huang, and Q.~Zou, ``{Unsupervised Part Mining for
  Fine-grained Image Classification},'' \emph{arXiv preprint arXiv:1902.09941},
  2019. [Online]. Available: \url{http://arxiv.org/abs/1902.09941}
\BIBentrySTDinterwordspacing

\bibitem{5}
R.~Ji, L.~Wen, L.~Zhang, D.~Du, Y.~Wu, C.~Zhao, X.~Liu, and F.~Huang,
  ``{Attention Convolutional Binary Neural Tree for Fine-Grained Visual
  Categorization},'' \emph{Proceedings of the IEEE Computer Society Conference
  on Computer Vision and Pattern Recognition}, pp. 10\,465--10\,474, 2020.

\bibitem{6}
W.~Luo, X.~Yang, X.~Mo, Y.~Lu, L.~Davis, J.~Li, J.~Yang, and S.~N. Lim,
  ``{Cross-X learning for fine-grained visual categorization},'' in
  \emph{Proceedings of the IEEE International Conference on Computer Vision},
  vol. 2019-Octob.\hskip 1em plus 0.5em minus 0.4em\relax Institute of
  Electrical and Electronics Engineers Inc., oct 2019, pp. 8241--8250.

\bibitem{8}
M.~Sun, Y.~Yuan, F.~Zhou, and E.~Ding, ``{Multi-Attention Multi-Class
  Constraint for Fine-grained Image Recognition},'' in \emph{Lecture Notes in
  Computer Science (including subseries Lecture Notes in Artificial
  Intelligence and Lecture Notes in Bioinformatics)}, vol. 11220 LNCS, 2018,
  pp. 834--850.

\bibitem{85}
Q.~Wang, J.~Wang, H.~Deng, X.~Wu, Y.~Wang, and G.~Hao, ``Aa-trans: Core
  attention aggregating transformer with information entropy selector for
  fine-grained visual classification,'' \emph{Pattern Recognition}, vol. 140,
  p. 109547, 2023.

\bibitem{45}
Y.~Wen, K.~Zhang, Z.~Li, and Y.~Qiao, ``{A discriminative feature learning
  approach for deep face recognition},'' in \emph{Lecture Notes in Computer
  Science (including subseries Lecture Notes in Artificial Intelligence and
  Lecture Notes in Bioinformatics)}, vol. 9911 LNCS.\hskip 1em plus 0.5em minus
  0.4em\relax Springer, 2016, pp. 499--515.

\bibitem{11}
P.~Du, Z.~Sun, Y.~Yao, and Z.~Tang, ``{Exploiting category similarity-based
  distributed labeling for fine-grained visual classification},'' \emph{IEEE
  Access}, vol.~8, pp. 186\,679--186\,690, 2020.

\bibitem{46}
A.~H. Farzaneh and X.~Qi, ``{Facial expression recognition in the wild via deep
  attentive center loss},'' \emph{Proceedings - 2021 IEEE Winter Conference on
  Applications of Computer Vision, WACV 2021}, pp. 2401--2410, 2021.

\bibitem{47}
J.~Li, H.~Xie, J.~Li, Z.~Wang, and Y.~Zhang, ``{Frequency-aware Discriminative
  Feature Learning Supervised by Single-Center Loss for Face Forgery
  Detection},'' Tech. Rep., 2021.

\bibitem{83}
Z.~Zhang, C.~Luo, H.~Wu, Y.~Chen, N.~Wang, and C.~Song, ``From individual to
  whole: reducing intra-class variance by feature aggregation,''
  \emph{International Journal of Computer Vision}, vol. 130, no.~3, pp.
  800--819, 2022.

\bibitem{10}
\BIBentryALTinterwordspacing
J.~He, J.-N. Chen, S.~Liu, A.~Kortylewski, C.~Yang, Y.~Bai, and C.~Wang,
  ``{TransFG: A Transformer Architecture for Fine-grained Recognition},''
  \emph{arXiv preprint arXiv:2103.07976}, 2021. [Online]. Available:
  \url{http://arxiv.org/abs/2103.07976}
\BIBentrySTDinterwordspacing

\bibitem{74}
R.~Hadsell, S.~Chopra, and Y.~LeCun, ``Dimensionality reduction by learning an
  invariant mapping,'' in \emph{2006 IEEE Computer Society Conference on
  Computer Vision and Pattern Recognition (CVPR'06)}, vol.~2.\hskip 1em plus
  0.5em minus 0.4em\relax IEEE, 2006, pp. 1735--1742.

\bibitem{75}
F.~Schroff, D.~Kalenichenko, and J.~Philbin, ``Facenet: A unified embedding for
  face recognition and clustering,'' in \emph{Proceedings of the IEEE
  conference on computer vision and pattern recognition}, 2015, pp. 815--823.

\bibitem{79}
Y.~Zeng, B.~Zhao, S.~Qiu, T.~Dai, and S.-T. Xia, ``Towards effective image
  manipulation detection with proposal contrastive learning,'' \emph{IEEE
  Transactions on Circuits and Systems for Video Technology}, 2023.

\bibitem{84}
S.~Zhang, J.~Bai, T.~Li, Z.~Yan, and Z.~Li, ``Modeling intra-class and
  inter-class constraints for out-of-domain detection,'' in \emph{International
  Conference on Database Systems for Advanced Applications}.\hskip 1em plus
  0.5em minus 0.4em\relax Springer, 2023, pp. 142--158.

\bibitem{72}
G.~Hinton, O.~Vinyals, J.~Dean \emph{et~al.}, ``Distilling the knowledge in a
  neural network,'' \emph{arXiv preprint arXiv:1503.02531}, vol.~2, no.~7,
  2015.

\bibitem{18}
C.~Szegedy, V.~Vanhoucke, S.~Ioffe, J.~Shlens, and Z.~Wojna, ``{Rethinking the
  Inception Architecture for Computer Vision},'' \emph{Proceedings of the IEEE
  Computer Society Conference on Computer Vision and Pattern Recognition}, vol.
  2016-Decem, pp. 2818--2826, 2016.

\bibitem{17}
C.~B. Zhang, P.~T. Jiang, Q.~Hou, Y.~Wei, Q.~Han, Z.~Li, and M.~M. Cheng,
  ``{Delving deep into label smoothing},'' \emph{IEEE Transactions on Image
  Processing}, vol.~30, pp. 5984--5996, 2021.

\bibitem{20}
\BIBentryALTinterwordspacing
H.~Bagherinezhad, M.~Horton, M.~Rastegari, and A.~Farhadi, ``{Label Refinery:
  Improving ImageNet Classification through Label Progression},'' \emph{arXiv
  preprint arXiv:1805.02641}, 2018. [Online]. Available:
  \url{http://arxiv.org/abs/1805.02641}
\BIBentrySTDinterwordspacing

\bibitem{73}
P.~Zhao, H.~Yao, X.~Liu, R.~Liu, and Q.~Miao, ``Improving image classification
  through joint guided learning,'' \emph{IEEE Transactions on Instrumentation
  and Measurement}, 2022.

\bibitem{21}
\BIBentryALTinterwordspacing
S.~Maji, E.~Rahtu, J.~Kannala, M.~Blaschko, and A.~Vedaldi, ``{Fine-Grained
  Visual Classification of Aircraft},'' 2013. [Online]. Available:
  \url{http://arxiv.org/abs/1306.5151}
\BIBentrySTDinterwordspacing

\bibitem{22}
B.~Englert and S.~Lam, ``{The Caltech-UCSD Birds-200-2011 Dataset},''
  \emph{IFAC Proceedings Volumes (IFAC-PapersOnline)}, vol.~42, no.~15, pp.
  50--57, 2009.

\bibitem{23}
J.~Krause, M.~Stark, J.~Deng, and L.~Fei-Fei, ``{3D object representations for
  fine-grained categorization},'' \emph{Proceedings of the IEEE International
  Conference on Computer Vision}, pp. 554--561, 2013.

\bibitem{91}
G.~Van~Horn, S.~Branson, R.~Farrell, S.~Haber, J.~Barry, P.~Ipeirotis,
  P.~Perona, and S.~Belongie, ``Building a bird recognition app and large scale
  dataset with citizen scientists: The fine print in fine-grained dataset
  collection,'' in \emph{Proceedings of the IEEE conference on computer vision
  and pattern recognition}, 2015, pp. 595--604.

\bibitem{24}
\BIBentryALTinterwordspacing
K.~He, X.~Zhang, S.~Ren, and J.~Sun, ``{Deep residual learning for image
  recognition},'' in \emph{Proceedings of the IEEE Computer Society Conference
  on Computer Vision and Pattern Recognition}, vol. 2016-Decem, 2016, pp.
  770--778. [Online]. Available:
  \url{http://openaccess.thecvf.com/content_cvpr_2016/html/He_Deep_Residual_Learning_CVPR_2016_paper.html}
\BIBentrySTDinterwordspacing

\bibitem{25}
\BIBentryALTinterwordspacing
A.~Dosovitskiy, L.~Beyer, A.~Kolesnikov, D.~Weissenborn, X.~Zhai,
  T.~Unterthiner, M.~Dehghani, M.~Minderer, G.~Heigold, S.~Gelly, J.~Uszkoreit,
  and N.~Houlsby, ``{An Image is Worth 16x16 Words: Transformers for Image
  Recognition at Scale},'' \emph{Icml}, pp. 1--21, 2020. [Online]. Available:
  \url{http://arxiv.org/abs/2010.11929}
\BIBentrySTDinterwordspacing

\bibitem{80}
J.~Li, L.~Yang, Q.~Wang, and Q.~Hu, ``Wdan: A weighted discriminative
  adversarial network with dual classifiers for fine-grained open-set domain
  adaptation,'' \emph{IEEE Transactions on Circuits and Systems for Video
  Technology}, 2023.

\bibitem{81}
T.~Yan, H.~Li, B.~Sun, Z.~Wang, and Z.~Luo, ``Discriminative feature mining and
  enhancement network for low-resolution fine-grained image recognition,''
  \emph{IEEE Transactions on Circuits and Systems for Video Technology},
  vol.~32, no.~8, pp. 5319--5330, 2022.

\bibitem{82}
S.~Wang, Z.~Wang, H.~Li, J.~Chang, W.~Ouyang, and Q.~Tian, ``Semantic-guided
  information alignment network for fine-grained image recognition,''
  \emph{IEEE Transactions on Circuits and Systems for Video Technology}, 2023.

\bibitem{33}
X.~S. Wei, C.~W. Xie, J.~Wu, and C.~Shen, ``{Mask-CNN: Localizing parts and
  selecting descriptors for fine-grained bird species categorization},''
  \emph{Pattern Recognition}, vol.~76, pp. 704--714, 2018.

\bibitem{34}
S.~Huang, Z.~Xu, D.~Tao, and Y.~Zhang, ``{Part-stacked CNN for fine-grained
  visual categorization},'' \emph{Proceedings of the IEEE Computer Society
  Conference on Computer Vision and Pattern Recognition}, vol. 2016-Decem, pp.
  1173--1182, 2016.

\bibitem{35}
N.~Zhang, J.~Donahue, R.~Girshick, and T.~Darrell, ``{Part-based R-CNNs for
  fine-grained category detection},'' \emph{Lecture Notes in Computer Science
  (including subseries Lecture Notes in Artificial Intelligence and Lecture
  Notes in Bioinformatics)}, vol. 8689 LNCS, no. PART 1, pp. 834--849, 2014.

\bibitem{36}
D.~Lin, X.~Shen, C.~Lu, and J.~Jia, ``{Deep LAC: Deep localization, alignment
  and classification for fine-grained recognition},'' \emph{Proceedings of the
  IEEE Computer Society Conference on Computer Vision and Pattern Recognition},
  vol. 07-12-June, pp. 1666--1674, 2015.

\bibitem{37}
S.~Branson, G.~{Van Horn}, S.~Belongie, and P.~Perona, ``{Bird species
  categorization using pose normalized deep convolutional nets},'' \emph{BMVC
  2014 - Proceedings of the British Machine Vision Conference 2014}, 2014.

\bibitem{86}
C.~Zhang, G.~Lin, Q.~Wang, F.~Shen, Y.~Yao, and Z.~Tang, ``Guided by meta-set:
  a data-driven method for fine-grained visual recognition,'' \emph{IEEE
  Transactions on Multimedia}, 2022.

\bibitem{87}
S.~Tsutsui, Y.~Fu, and D.~Crandall, ``Reinforcing generated images via
  meta-learning for one-shot fine-grained visual recognition,'' \emph{IEEE
  Transactions on Pattern Analysis and Machine Intelligence}, 2022.

\bibitem{94}
Z.~Yang, T.~Luo, D.~Wang, Z.~Hu, J.~Gao, and L.~Wang, ``Learning to navigate
  for fine-grained classification,'' in \emph{Proceedings of the European
  conference on computer vision (ECCV)}, 2018, pp. 420--435.

\bibitem{93}
X.~He, Y.~Peng, and J.~Zhao, ``Which and how many regions to gaze: Focus
  discriminative regions for fine-grained visual categorization,''
  \emph{International Journal of Computer Vision}, vol. 127, pp. 1235--1255,
  2019.

\bibitem{92}
H.~Sun, X.~He, and Y.~Peng, ``Sim-trans: Structure information modeling
  transformer for fine-grained visual categorization,'' in \emph{Proceedings of
  the 30th ACM International Conference on Multimedia}, 2022, pp. 5853--5861.

\bibitem{42}
R.~Du, D.~Chang, A.~K. Bhunia, J.~Xie, Z.~Ma, Y.~Z. Song, and J.~Guo,
  ``{Fine-Grained Visual Classification via Progressive Multi-granularity
  Training of Jigsaw Patches},'' \emph{Lecture Notes in Computer Science
  (including subseries Lecture Notes in Artificial Intelligence and Lecture
  Notes in Bioinformatics)}, vol. 12365 LNCS, pp. 153--168, 2020.

\bibitem{43}
P.~Zhao, Q.~Miao, H.~Yao, X.~Liu, R.~Liu, and M.~Gong, ``{CA-PMG: Channel
  attention and progressive multi-granularity training network for fine-grained
  visual classification},'' \emph{IET Image Processing}, vol.~15, no.~14, pp.
  3718--3727, 2021.

\bibitem{44}
Y.~Chen, Y.~Bai, W.~Zhang, and T.~Mei, ``{Destruction and construction learning
  for fine-grained image recognition},'' in \emph{Proceedings of the IEEE
  Computer Society Conference on Computer Vision and Pattern Recognition}, vol.
  2019-June, 2019, pp. 5152--5161.

\bibitem{77}
R.~Ji, J.~Li, L.~Zhang, J.~Liu, and Y.~Wu, ``Dual transformer with
  multi-grained assembly for fine-grained visual classification,'' \emph{IEEE
  Transactions on Circuits and Systems for Video Technology}, 2023.

\bibitem{78}
J.~Peng, G.~Jiang, and H.~Wang, ``Adaptive memorization with group labels for
  unsupervised person re-identification,'' \emph{IEEE Transactions on Circuits
  and Systems for Video Technology}, 2023.

\bibitem{88}
H.~Huang, Z.~Wu, W.~Li, J.~Huo, and Y.~Gao, ``Local descriptor-based
  multi-prototype network for few-shot learning,'' \emph{Pattern Recognition},
  vol. 116, p. 107935, 2021.

\bibitem{89}
\BIBentryALTinterwordspacing
J.~Snell, K.~Swersky, and R.~Zemel, ``Prototypical networks for few-shot
  learning,'' in \emph{Advances in Neural Information Processing Systems},
  I.~Guyon, U.~V. Luxburg, S.~Bengio, H.~Wallach, R.~Fergus, S.~Vishwanathan,
  and R.~Garnett, Eds., vol.~30.\hskip 1em plus 0.5em minus 0.4em\relax Curran
  Associates, Inc., 2017. [Online]. Available:
  \url{https://proceedings.neurips.cc/paper_files/paper/2017/file/cb8da6767461f2812ae4290eac7cbc42-Paper.pdf}
\BIBentrySTDinterwordspacing

\bibitem{90}
H.~Chen, H.~Li, Y.~Li, and C.~Chen, ``Sparse spatial transformers for few-shot
  learning,'' \emph{arXiv preprint arXiv:2109.12932}, 2021.

\bibitem{48}
\BIBentryALTinterwordspacing
T.~Miyato, S.~I. Maeda, M.~Koyama, K.~Nakae, and S.~Ishii, ``{Distributional
  Smoothing with Virtual Adversarial Training},'' \emph{4th International
  Conference on Learning Representations, ICLR 2016 - Conference Track
  Proceedings}, 2018. [Online]. Available:
  \url{http://www.shortscience.org/paper?bibtexKey=journals/corr/1507.00677#davidstutz}
\BIBentrySTDinterwordspacing

\bibitem{95}
``{iNaturalist} 2018 competition dataset.''
  ~\url{https://github.com/visipedia/inat_comp/tree/master/2018}, 2018.

\bibitem{15}
A.~Dubey, O.~Gupta, P.~Guo, R.~Raskar, R.~Farrell, and N.~Naik, ``{Pairwise
  confusion for fine-grained visual classification},'' \emph{Lecture Notes in
  Computer Science (including subseries Lecture Notes in Artificial
  Intelligence and Lecture Notes in Bioinformatics)}, vol. 11216 LNCS, pp.
  71--88, 2018.

\bibitem{59}
G.~Huang, Z.~Liu, L.~Van Der~Maaten, and K.~Q. Weinberger, ``Densely connected
  convolutional networks,'' in \emph{Proceedings of the IEEE conference on
  computer vision and pattern recognition}, 2017, pp. 4700--4708.

\bibitem{50}
\BIBentryALTinterwordspacing
Z.~Liu, Y.~Lin, Y.~Cao, H.~Hu, Y.~Wei, Z.~Zhang, S.~Lin, and B.~Guo, ``{Swin
  Transformer: Hierarchical Vision Transformer using Shifted Windows},''
  \emph{2021 IEEE/CVF International Conference on Computer Vision (ICCV)}, pp.
  9992--10\,002, mar 2021. [Online]. Available:
  \url{http://arxiv.org/abs/2103.14030
  https://ieeexplore.ieee.org/document/9710580/}
\BIBentrySTDinterwordspacing

\bibitem{49}
\BIBentryALTinterwordspacing
Y.~Rao, G.~Chen, J.~Lu, and J.~Zhou, ``{Counterfactual Attention Learning for
  Fine-Grained Visual Categorization and Re-identification},'' in
  \emph{Proceedings of the IEEE/CVF International Conference on Computer
  Vision}, 2021, pp. 1025--1034. [Online]. Available:
  \url{http://arxiv.org/abs/2108.08728}
\BIBentrySTDinterwordspacing

\bibitem{51}
T.~Y. Lin, A.~Roychowdhury, and S.~Maji, ``{Bilinear Convolutional Neural
  Networks for Fine-Grained Visual Recognition},'' \emph{IEEE Transactions on
  Pattern Analysis and Machine Intelligence}, vol.~40, no.~6, pp. 1309--1322,
  2018.

\bibitem{57}
Y.~Liang, L.~Zhu, X.~Wang, and Y.~Yang, ``{A Simple Episodic Linear Probe
  Improves Visual Recognition in the Wild},'' in \emph{Proceedings of the
  IEEE/CVF Conference on Computer Vision and Pattern Recognition}, 2022, pp.
  9559--9569.

\bibitem{70}
P.~Zhuang, Y.~Wang, and Y.~Qiao, ``{Learning attentive pairwise interaction for
  fine-grained classification},'' in \emph{AAAI 2020 - 34th AAAI Conference on
  Artificial Intelligence}, 2020, pp. 13\,130--13\,137.

\bibitem{61}
C.~Jia, Y.~Yang, Y.~Xia, Y.~T. Chen, and T.~Duerig, ``Scaling up visual and
  vision-language representation learning with noisy text supervision,'' 2021.

\bibitem{69}
\BIBentryALTinterwordspacing
H.~Zhu, W.~Ke, D.~Li, J.~Liu, L.~Tian, and Y.~Shan, ``{Dual Cross-Attention
  Learning for Fine-Grained Visual Categorization and Object
  Re-Identification},'' 2022. [Online]. Available:
  \url{http://arxiv.org/abs/2205.02151}
\BIBentrySTDinterwordspacing

\bibitem{54}
Z.~Miao, X.~Zhao, J.~Wang, Y.~Li, and H.~Li, ``{Complemental Attention
  Multi-Feature Fusion Network for Fine-Grained Classification},'' \emph{IEEE
  Signal Processing Letters}, vol.~28, pp. 1983--1987, 2021.

\bibitem{55}
\BIBentryALTinterwordspacing
L.~van~der Maaten and G.~Hinton, ``{Visualizing data using t-SNE},''
  \emph{Journal of Machine Learning Research}, vol.~9, no.~1, pp. 2579--2605,
  2008. [Online]. Available:
  \url{https://www.jmlr.org/papers/volume9/vandermaaten08a/vandermaaten08a.pdf?fbclid=IwA}
\BIBentrySTDinterwordspacing

\bibitem{71}
C.~Qi and F.~Su, ``{Contrastive-center loss for deep neural networks},''
  \emph{Proceedings - International Conference on Image Processing, ICIP}, vol.
  2017-Septe, pp. 2851--2855, 2018.

\bibitem{56}
\BIBentryALTinterwordspacing
R.~R. Selvaraju, M.~Cogswell, A.~Das, R.~Vedantam, D.~Parikh, and D.~Batra,
  ``{Grad-cam: Why did you say that? visual explanations from deep networks via
  gradient-based localization},'' \emph{Revista do Hospital das Cl??nicas},
  vol.~17, pp. 331--336, 2016. [Online]. Available:
  \url{http://arxiv.org/abs/1610.02391}
\BIBentrySTDinterwordspacing

\end{thebibliography}

\newpage
\appendix
\section{Experiment on the large-scale dataset}

\IEEEPARstart{T}{o} order to verify the effectiveness of the proposed ECC on the large-scale dataset, we conduct experiments on the iNaturalist 2018 dataset (iNat2018), which includes 8142 species, 437513 training images and 24426 validation images. There is a serious long-tail problem: the category with the most samples in the training set has several thousand samples, while the category with the least samples has only a few samples. 

In the experiments, we use ResNet50 as backbone. $\lambda_1$ is set as 0.05, and $\lambda_2$ is set as 0.001. The other hyperparameters remain consistent with those of the other datasets (such as batch size is 32 and initial learning is 0.01). In addition, we do not use additional information, including latitude, longitude and date.

We explore the efficiency of the proposed ECC and two components (the MCC and CLG). The results are shown in \textcolor{blue}{Table.\ref{table:tables1}}. Although the main challenge of the iNat2018 dataset is the long-tail problem, rather than intra-class variances and inter-classes differences, our approach still brings improvements.

\begin{table}
	\caption{Results of our method ECC and two components (MCC and CLG) on iNat2018 dataset. The best results are shown in bold.}
	\centering
	\label{table:tables1}
	\begin{tabular}{c|c|c}
		\hline
		\textbf{Components} & \textbf{Backbone} & \textbf{iNat2018} \\
		\hline
		Baseline &ResNet50 &62.4\\
		MCC &ResNet50 &62.7 \\
		CLG &ResNet50 &62.5 \\
		ECC (MCC+CLG) &ResNet50 &\textbf{62.8}\\
		\hline
	\end{tabular}
\end{table}
\end{document}